\documentclass[sigconf]{acmart}
\AtBeginDocument{%
  }

\setcopyright{acmlicensed}
\copyrightyear{2018}
\acmYear{2018}
\acmDOI{XXXXXXX.XXXXXXX}
\acmConference[Conference acronym 'XX]{Make sure to enter the correct
  conference title from your rights confirmation email}{June 03--05,
  2018}{Woodstock, NY}
\acmISBN{978-1-4503-XXXX-X/2018/06}

\usepackage{times}
\usepackage{url}
\usepackage{soul}
\usepackage{multirow}
\usepackage{lipsum} 
\usepackage{booktabs}
\usepackage{multirow}
\usepackage{enumitem}
\usepackage{subcaption}
\usepackage{array}
\usepackage{tcolorbox}
\usepackage{amsthm}
\usepackage{booktabs}
\usepackage{algorithm}
\usepackage{algorithmic}
\usepackage[switch]{lineno}
\usepackage{tabularx}
\usepackage{tikz}
\usepackage{soul}
\usepackage{float}    
\usepackage{colortbl}
\usepackage{xcolor}
\newcommand{\methodname}{H-FedSN}
\begin{document}

\title{H-FedSN: Personalized Sparse Networks for Efficient and Accurate Hierarchical Federated Learning for IoT Applications}

%
% The "author" command and its associated commands are used to define
% the authors and their affiliations.
% Of note is the shared affiliation of the first two authors, and the
% "authornote" and "authornotemark" commands
% used to denote shared contribution to the research.

% \author{\IEEEauthorblockN{Jiechao Gao\textsuperscript{*}}
% \IEEEauthorblockA{\textit{Department of Computer Science} \\
% \textit{University of Virginia}\\
% Charlottesville, VA, USA\\
% jg5ycn@virginia.edu}
% \and
% \IEEEauthorblockN{Yuangang Li\textsuperscript{*}}
% \IEEEauthorblockA{\textit{Thomas Lord Department of Computer Science} \\
% \textit{University of Southern California}\\
% Los Angeles, CA, USA \\
% yuangang@usc.edu}
% \and
% \IEEEauthorblockN{ Yue Zhao\textsuperscript{\dag}}
% \IEEEauthorblockA{\textit{Thomas Lord Department of Computer Science} \\
% \textit{University of Southern California}\\
% Los Angeles, CA, USA \\
% yzhao010@usc.edu}
% \and
% \IEEEauthorblockN{Brad Campbell\textsuperscript{\dag}}
% \IEEEauthorblockA{\textit{Department of Computer Science} \\
% \textit{University of Virginia}\\
% Charlottesville, VA, USA\\
% bradjc@virginia.edu}
% \thanks{\textsuperscript{*}Equal contribution.}
% \thanks{\textsuperscript{\dag}Corresponding authors.}
% }

\author{Jiechao Gao}
\authornote{Both authors contributed equally to this research.}
\affiliation{%
  \institution{Stanford University}
  \city{Stanford}
  \state{CA}
  \country{USA}
}
\email{jiechao@stanford.edu}

\author{Yuangang Li\textsuperscript{*}}
\affiliation{%
  \institution{University of California, Irvine}
  \city{Irvine}
  \state{CA}
  \country{USA}
}
\email{yuanganl@uci.edu}

\author{Jie Wang}
\affiliation{%
  \institution{Stanford University}
  \city{Stanford}
  \state{CA}
  \country{USA}
}
\email{jiewang@stanford.edu}

\author{Michael Lepech}
\affiliation{%
  \institution{Stanford University}
  \city{Stanford}
  \state{CA}
  \country{USA}
}
\email{mlepech@stanford.edu}

\author{Yue Zhao}
\affiliation{%
 \institution{University of Southern California}
 \city{Los Angeles}
 \state{CA}
 \country{USA}
}
\email{yue.z@usc.edu}

\author{Brad Campbell}
\affiliation{%
  \institution{University of Virginia}
  \city{Charlottesville}
  \state{VA}
  \country{USA}
}
\email{bradjc@virginia.edu}

% By default, the full list of authors will be used in the page
% headers. Often, this list is too long, and will overlap
% other information printed in the page headers. This command allows
% the author to define a more concise list
% of authors' names for this purpose.
\renewcommand{\shortauthors}{Gao et al.}

\begin{abstract}
With the rapid development of the Internet of Things (IoT), federated learning (FL) has gained increasing attention for its privacy-preserving use of distributed data. However, conventional two-tier FL architectures are poorly suited to the hierarchical and heterogeneous nature of real-world IoT systems. Hierarchical Federated Learning (HFL) introduces multi-layer aggregation to better match IoT environments, but still suffers from communication inefficiencies and performance limitations caused by large data transfers, non-IID data distributions, and uneven device participation. These challenges hinder the realization of low-latency and high-accuracy training in practical IoT deployments.
To address these limitations, we propose H-FedSN for practical IoT environments. H-FedSN leverages a binary mask mechanism with shared and personalized layers to reduce communication overhead by creating a sparse network without altering original weights. To tackle data heterogeneity and imbalanced device distribution, H-FedSN incorporates personalized layers for local data adaptation and employs Bayesian aggregation with cumulative Beta distribution updates at edge and cloud levels, effectively balancing contributions from diverse client groups.
Experiments on three real-world IoT datasets and MNIST under non-IID conditions show that H-FedSN reduces communication costs by up to 477 times compared to baseline methods while maintaining high accuracy, making it well-suited for hierarchical FL in practical IoT deployments.
\end{abstract}



% \keywords{Do, Not, Use, This, Code, Put, the, Correct, Terms, for,

\maketitle

\section{Introduction} 
Federated learning (FL) enables multiple distributed devices to collaboratively train a model without sharing raw data, allowing global knowledge integration while preserving data privacy~\cite{yang2019federated,mcmahan2017communication}. 
Integrating FL into ubiquitous Internet of Things (IoT) environments unlocks a wide range of applications~\cite{khan2021federated,zhang2022federated}. 
As IoT systems continue to expand in scale and complexity, the data collection, computing, and communication resources follows a natural multi-tier structure\cite{shi2016edge,satyanarayanan2017emergence}. 
Terminal devices operate at the sensing layer and continually generate data; edge servers provide intermediate computation and aggregation; and cloud servers execute global coordination and optimization. 
For example, in smart city surveillance, cameras connect to building-level servers before interacting with the cloud~\cite{anthopoulos2015understanding}; in smart agriculture, drones and ground sensors upload data to farm-level gateways that then communicate with regional servers~\cite{Virk2020Smart}; and in healthcare monitoring, wearable devices transmit physiological signals to nearby edge gateways before being forwarded to the cloud~\cite{quy2022smart}.
Given this intrinsic layering of IoT systems, collaborative learning frameworks generally follow the corresponding collaboration pathways among terminals, edges, and the cloud.

\begin{figure}[t!]
\centering
\includegraphics[width=\linewidth]{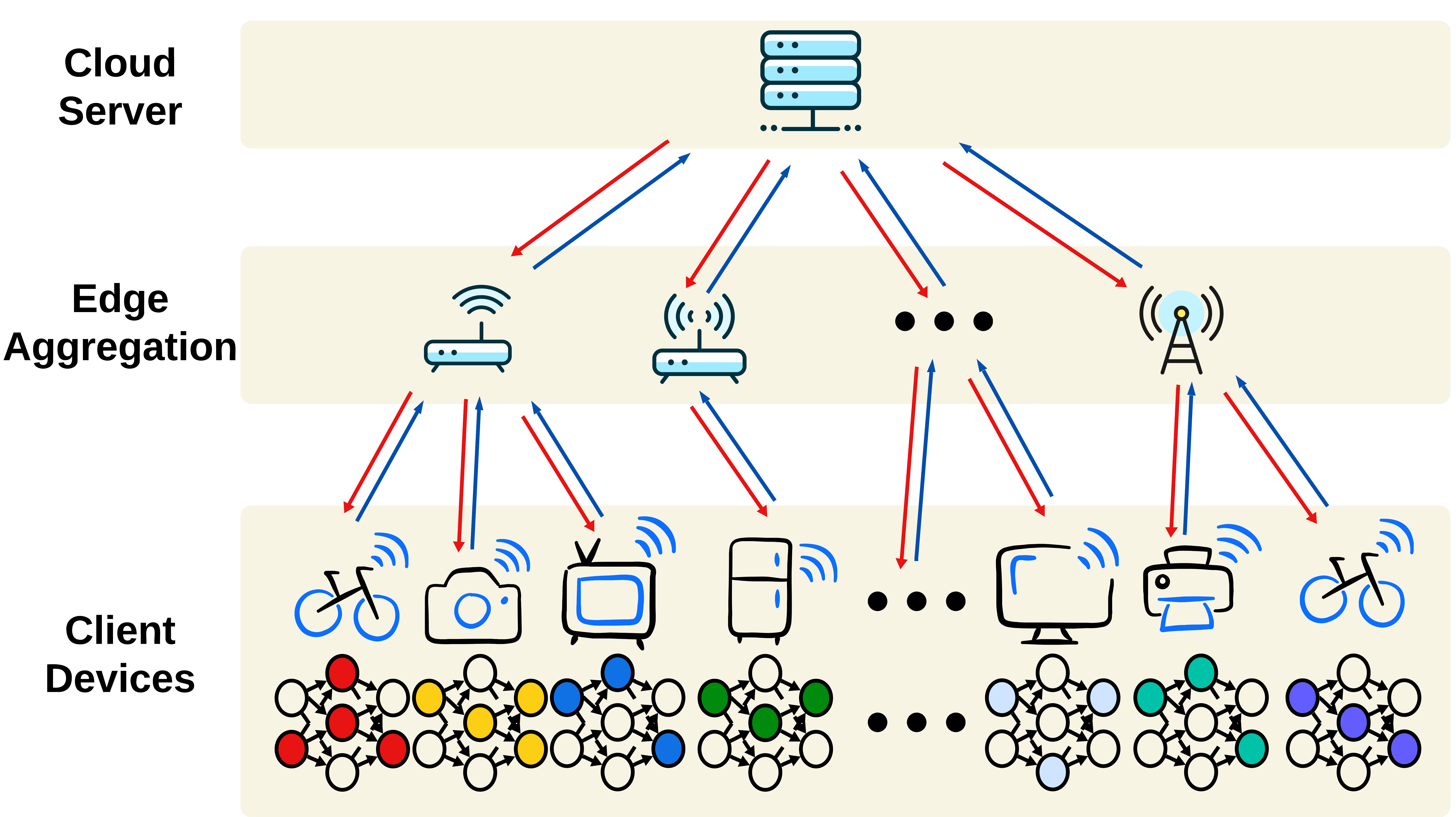} 
\caption{\methodname{}: Personalized and structured sparse models for each client to achieve high communication efficiency while
maintaining model accuracy within the HFL architecture for realistic IoT environment.}
\label{fig_alg_highlevel}
\end{figure}

Compared to two-tier federated learning, hierarchical federated learning (HFL)\cite{liu2020client} is inherently more suitable for multi-tier architectures of IoT systems and is thus more practical for real-world deployments. However, it still faces key challenges in communication efficiency and model accuracy.
In particular, bandwidth constraints and communication latency pose critical bottlenecks that degrade HFL’s training efficiency and convergence speed across tiers. Limited bandwidth restricts the transmission of model parameters, prolonging training rounds, while high latency introduces cascading delays during hierarchical aggregation. For example, HierFAVG~\cite{liu2020client} requires devices to transmit full model parameters over bandwidth-limited IoT links across multiple layers, making real-world deployment impractical.
Additionally, IoT devices typically encounter non-independent and identically distributed (non-IID) data and imbalanced client distributions, which further degrade model accuracy~\cite{9533879}.
Therefore, designing communication-efficient HFL strategies that mitigate latency and bandwidth limitations while maintaining high model accuracy is essential to enable scalable and high-performance IoT applications~\cite{Sheng2023Federated}.

Although many approaches~\cite{liu2024fedbcgd,10335537EAFL,9288933FTTQ} have been proposed to enhance communication efficiency in conventional two-tier federated learning, adapting these methods to IoT scenarios, particularly in three-tier hierarchical federated learning (HFL), remains challenging due to limited research on architectural adaptation and the risk of performance degradation. For example, quantization-based techniques~\cite{9288933FTTQ} may accumulate errors across multiple aggregation layers, significantly impairing model accuracy. In contrast, personalized federated learning algorithms~\cite{10.1145/3447548.3467254,fallah2020personalized,arivazhagan2019federated} are well-suited to address data heterogeneity and can be integrated into HFL frameworks with relative ease, offering improved accuracy. However, these methods typically neglect communication efficiency. HFL-based methods also face limitations in balancing accuracy and communication efficiency, such as HierFAVG~\cite{liu2020client} that, despite fitting hierarchical IoT architectures, transmits full model parameters incurring high communication costs, and ESPerHFL~\cite{Ma_2024} that, although achieving better accuracy via personalization, introduces additional parameters thus increasing communication overhead.

To balance communication costs and accuracy, we propose \methodname{}, designed for real-world IoT environments to achieve low communication costs and high accuracy in the HFL architecture.
\methodname{} centers on a personalized, structured sparse model for each client, as shown in Fig.~\ref{fig_alg_highlevel}.
Clients start with an identical neural network with frozen weights and train a corresponding binary (0,1) mask network. This binary mask network has the same structure as the original neural network. 
Applying this binary mask to the original network determines which connections to retain or prune, thereby creating a sparse network. 
The mask network consists of shared layers and personalized layers. Shared layers participate in global aggregation, becoming identical across all clients after training. 
Personalized layers are trained on client-specific data, remain local, and do not participate in global aggregation, resulting in unique layers for each client.
To further enhance model accuracy and address imbalanced client
distribution, \methodname{} incorporates Bayesian aggregation at both edge and cloud levels. This method uses cumulative updates of Beta distribution parameters to balance contributions from edge nodes with varying numbers of connected clients. While edge nodes with more clients have a larger immediate impact on parameter updates, the cumulative nature of these updates ensures a fair representation of all edge nodes over time. Combined with the locally retained personalized mask layers, this approach allows the system to adapt to imbalanced client distributions across edges and achieve high model accuracy.

Experiments on three real-world IoT datasets
and MNIST under non-IID conditions, we compare our \methodname{} with 6 baselines, show that the communication cost of \methodname{} was lower than all baseline methods, reducing it by up to 477 times compared to baseline methods. Regarding accuracy, \methodname{} achieves much better results by an average of 8.6\% compared to communication-enhanced FL methods and is comparable to personalized FL methods within HFL architecture. 

\section{Related Work}
\label{app:detailed_hfl_review}
\subsection{Federated Learning and Its Limitations in Real-World IoT Scenarios}

FL has revolutionized the approach to distributed machine learning by enabling multiple client nodes to collaboratively train models without the need to centralize data, thereby enhancing privacy and reducing the risks of data breaches. Pioneering algorithms such as FedAvg \cite{mcmahan2017communication}, FedProx \cite{li2020federated}, and Scaffold \cite{karimireddy2020scaffold} have set the foundation for FL by demonstrating how collaborative training can be effectively managed across diverse nodes.

However, these traditional FL approaches face significant challenges when applied to real-world Internet of Things (IoT) scenarios. One of the primary limitations of these conventional FL algorithms is their reliance on a central server for model aggregation. This architecture poses several challenges: Scalability issues arise as the number of IoT devices grows, causing the central server’s capacity to handle simultaneous model updates to diminish, which can lead to processing delays and aggregation bottlenecks when dealing with thousands or even millions of devices \cite{Imteaj2021A}. Additionally, the geographic dispersion of IoT devices exacerbates communication delays and increases the risk of network failures; devices located far from the central server may experience significant lag, affecting the synchronization and efficiency of the FL process \cite{Ni2023Semi-Federated}. Moreover, IoT devices often have limited computational and energy resources, and frequent communication with a distant central server can quickly deplete these resources, making FL impractical for many IoT applications \cite{Khan2020Federated}. Furthermore, the central server represents a single point of failure in the network; if the server goes offline or experiences issues, the entire federated learning process can be disrupted, leading to potential data loss and service interruptions \cite{Wu2020Personalized}.

\subsection{Hierarchical Federated Learning (HFL)}
To address the scalability issues inherent in traditional FL, Hierarchical Federated Learning (HFL) has been proposed as a robust solution. HFL introduces a multi-tier architecture, typically consisting of client, edge, and cloud layers. Liu et al. introduced the HierFAVG algorithm, which organizes clients into a hierarchical, tree-like topology~\cite{liu2020client}. The hierarchical approach of HFL is particularly suited to complex and geographically distributed environments, such as those found in IoT systems. By utilizing a multi-level communication structure, HFL can more effectively manage the diverse and distributed nature of IoT devices. However, this architecture also brings new challenges, particularly in terms of communication efficiency and coordination across different layers of the hierarchy \cite{Sheng2023Federated}.

\subsection{Communication Efficiency in FL and HFL}
In the traditional FL, various algorithms have been developed to address communication issues and enhance communication efficiency. For instance, FedSCR introduces a structure-based communication reduction method, reducing upstream communication by nearly 50\%, but this comes with a tolerable accuracy drop \cite{9303442}. Similarly, Fast Federated Learning by Balancing Communication Trade-Offs reduces communication costs by balancing trade-offs between communication and computation, though this approach also results in some accuracy degradation \cite{Nori2021Fast}. Another example is the approach detailed in “Toward Communication-Learning Trade-Off for Federated Learning at the Network Edge,” which reduces communication overhead through network pruning, but this technique leads to accuracy degradation due to information loss \cite{Ren2022Toward}. While these algorithms have made significant progress in improving communication efficiency, they all exhibit some level of accuracy reduction.

Despite these advancements, research specifically focusing on communication strategies within HFL remains sparse. The unique multi-level communication requirements in HFL complicate the straightforward application of traditional FL techniques. Consequently, there is a significant gap in the literature regarding the optimization of communication within HFL. Existing communication algorithms developed for FL often lack the adaptability required for HFL’s multi-level structure, leading to performance trade-offs and reduced effectiveness.

We propose a novel algorithm designed specifically for HFL, named \methodname{}. Our approach introduces network sparsification through a dual-layer masking mechanism applied at the client level. Each client initially trains a mask network to sparsify the model, significantly reducing the communication cost by transmitting only the binary mask instead of the entire model weights. This method maintains model accuracy while significantly lowering communication overhead. Additionally, \methodname{} incorporates personalized layers to handle unique client data characteristics and employs Bayesian aggregation for more robust and efficient model updates.

\section{\methodname{} Design}
\subsection{Overview}
\begin{figure}[htbp]
\centering
\includegraphics[width=\linewidth]{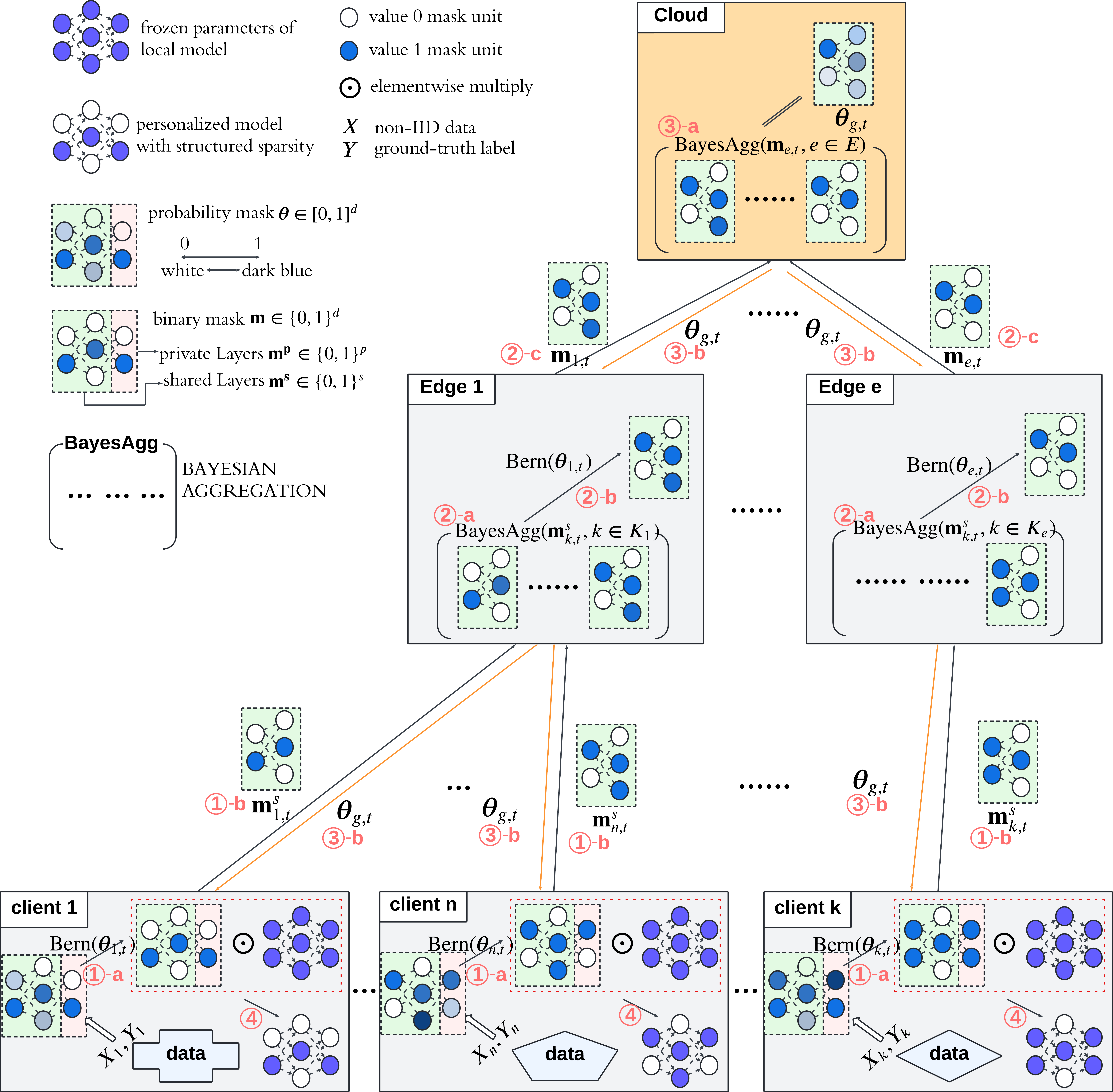}
\caption{Overview of \methodname{}  framework.}
\label{fig_alg}
\end{figure}

\noindent
\methodname{} is an algorithm based on the HFL architecture, designed to optimize communication efficiency and ensure model accuracy through efficient mask training, personalized layers, and Bayesian aggregation.
Zhou et al. demonstrated that training sparse models with masks, while keeping the model parameters frozen, is an effective alternative to traditional training methods \cite{zhou2019deconstructing}.

\textbf{Problem formulation.} We consider a three-tier hierarchical federated learning system with $|K|$ clients, $|E|$ edge servers, and a single cloud server. Each edge $e \in E$ serves a disjoint subset of clients $K_e \subseteq K$ (with $K = \bigcup_{e \in E} K_e$). Each client $k$ owns a private, potentially non-IID dataset $D_k$ that never leaves the device. Given a frozen shared initialization $\mathbf{w}_{\text{init}} \in \mathbb{R}^d$, each client $k$ trains a personalized probability mask $\boldsymbol{\theta}_k \in [0,1]^d$. Its stochastic binarization $\mathbf{m}_k \sim \mathrm{Bern}(\boldsymbol{\theta}_k) \in \{0,1\}^d$ decomposes as $\mathbf{m}_k = (\mathbf{m}_k^s, \mathbf{m}_k^p)$ over shared and client-private layers, and induces a sparse model $\dot{w}_k = \mathbf{m}_k \odot \mathbf{w}_{\text{init}}$. The system jointly trains the per-client masks $\{\boldsymbol{\theta}_k\}_{k \in K}$ to minimize each client's expected local task loss on its private data $D_k$, subject to three communication and privacy constraints: (i) raw data $D_k$ never leaves client $k$; (ii) each client $k \in K_e$ uploads only the shared binary mask $\mathbf{m}_k^s$ to its edge server $e$, and each edge $e$ uploads only its aggregated edge mask to the cloud; (iii) the private mask $\mathbf{m}_k^p$ is retained on client $k$ throughout training and never aggregated.

Fig.~\ref{fig_alg} depicts an overview of the proposed \methodname{} framework.
\methodname{} allows each client to freeze its random initialized local model parameters $\mathbf{w}_{\text{init}} = (w_{\text{init}}^1, w_{\text{init}}^2, \ldots, w_{\text{init}}^d) \in \mathbb{R}^d$ while collaboratively training a personalized probability mask $\boldsymbol{\theta}_k \in [0,1]^d$ for each client $k$, using their respective local datasets \(D_k\). This probability mask $\boldsymbol{\theta}_k$ determines the activation probability of the local model parameters and serves as the Bernoulli parameter for the stochastic binary mask $\mathbf{m}_k \sim \textbf{Bern}(\boldsymbol{\theta}_k) \in \{0,1\}^d$ (\textcircled{1}-a). The binary mask $\mathbf{m}_k$ is then applied to the frozen parameters $\mathbf{w}_{\text{init}}$ to create a structured sparse model, uniquely personalized for each client (\textcircled{4}). The resulting model, expressed as $\dot{w}_k = \mathbf{m}_k \odot \mathbf{w}_{\text{init}}$, is optimized to enhance performance on specific tasks. For a more detailed description of the algorithm, refer to Algorithm~\ref{alg:H-FedSN}.
\begin{algorithm}[t!]
\begin{algorithmic}[1]

\REQUIRE Number of training rounds $T$, number of local iterations $\tau$, learning rate $\eta$, initial Beta distribution parameters $\lambda_0 = 1$
\ENSURE Training a personalized and structured sparse model unique to each client.
\STATE At the cloud server, initialize Beta priors $\alpha_g^0 = \beta_g^0 = \lambda_0$,  and initialize a random network with weight vector  $\mathbf{w}_{\text{init}}\in \mathbb{R}^d$, and then broadcast it to the clients through edges.
\STATE At the edge servers, initialize Beta priors $\alpha_e^0 = \beta_e^0 = \lambda_0$.
\ \STATE For all clients, initialize random mask probability \(\boldsymbol{s}_{k,1}\)

\FOR{$t = 1$ to $T$}
    \FOR{each edge server $e \in E$ \textbf{in parallel}}
        \STATE \textbf{Edge server:}
        \FOR{each client $k \in K_e$ \textbf{in parallel}}
            \STATE \textbf{Client:}
             \IF {\(t > 1\)}
                \STATE $\boldsymbol{\theta}_{k,t} \gets \boldsymbol{\theta}_{g,t-1} + \boldsymbol{\theta}_{k,t-1}^p$
                \STATE $\mathbf{s}_{k,t} \gets \text{Sigmoid}^{-1}(\boldsymbol{\theta}_{k,t})$
            \ENDIF
            \FOR{$i = 1$ to $\tau$}
                \STATE $\boldsymbol{\theta}_{k,t} \gets \text{Sigmoid}(\mathbf{s}_{k,t})$
                \STATE $\mathbf{m}_{k,t} \gets \textbf{Bern}(\boldsymbol{\theta}_{k,t})$
                \STATE $\dot{w}_{k,t} \gets \mathbf{m}_{k,t} \odot \mathbf{w}_{\text{init}}$
                \STATE Compute loss $L(f(\dot{w}_{k,t}), D_k)$
                \STATE $\mathbf{s}_{k,t} \gets \mathbf{s}_{k,t} - \eta \nabla L(f(\dot{w}_{k,t}), D_k)$
            \ENDFOR
            
            \STATE $\boldsymbol{\theta}_{k,t} \gets \text{Sigmoid}(\mathbf{s}_{k,t})$
            \STATE $\mathbf{m}_{k,t} \gets \textbf{Bern}(\boldsymbol{\theta}_{k,t})$
            \STATE Decompose $\mathbf{m}_{k,t}$ into shared $\mathbf{m}_{k,t}^s$ (over shared layers) and private $\mathbf{m}_{k,t}^p$ (over personalized layers).
            \STATE Upload $\mathbf{m}_{k,t}^s$ to edge server $e$
        \ENDFOR
        
        \STATE \textbf{Edge server:}
        \STATE $\mathbf{m}_{e,t} = \text{BayesianAggregationAtEdge}(\{\mathbf{m}_{k,t}^s\}_{k \in K_e}, t)$
        \STATE Upload $\mathbf{m}_{e,t}$ to cloud server
    \ENDFOR
    
    \STATE \textbf{Cloud server:}
    \STATE $\boldsymbol{\theta}_{g,t} = \text{BayesianAggregationAtCloud}(\{\mathbf{m}_{e,t}\}_{e \in E},t)$
    \STATE Broadcast $\boldsymbol{\theta}_{g,t}$ to all edges $e \in E$, and all clients
\ENDFOR
\STATE For all clients,  $\mathbf{m}_{k,final} = \textbf{Bern}(\boldsymbol{\theta}_{g,T} + \boldsymbol{\theta}_{k,T}^p)$.
Generate the final model: $\dot{w}_{k,final} \gets \mathbf{m}_{k,final} \odot \mathbf{w}_{\text{init}}$.

\caption{\methodname{}}
\label{alg:H-FedSN}
\end{algorithmic}
\end{algorithm}

At the client level, each client $k$ is associated with an edge server $e$ and is part of the set $K_e$. Here, $e$ represents a specific edge server, and $K_e$ denotes the set of all clients connected to that particular edge server $e$. This setup implies multiple edge servers, each managing its own group of clients. Clients are responsible for training a probability mask $\boldsymbol{\theta}_k$, which is dimensioned $[0,1]^d$ to match the size of the client's local model parameters. Especially, the probability mask is bifurcated into a shared layer $\boldsymbol{\theta}_k^s \in [0,1]^s$ and a private layer $\boldsymbol{\theta}_k^p\in [0,1]^p$. The shared layer's parameters, meant for aggregation, enhance the model by leveraging collective insights, while the private layer's parameters are tailored for local data specificity and privacy, hence not shared. 
Based on this probability mask, the binary mask $\mathbf{m}_k = \text{Bern}(\boldsymbol{\theta}_k)$ inherits the bifurcated structure of $\boldsymbol{\theta}_k$, comprising shared $\mathbf{m}_k^s$ and private $\mathbf{m}_k^p$ layers. This structured approach ensures that while personalization is maintained through private layers, the collaborative benefits of federated learning are realized by the shared layers. 

Each edge server $e$, where $e \in E$, is responsible for collecting binary masks $\mathbf{m}_k^s$, $k \in K_e$  (\textcircled{1}-b). These edge servers aggregate (\textcircled{2}-a) the masks using a Bayesian aggregation method, which allows adaptation to various data distributions and improves performance through refined model updates. Post-aggregation, the edge server computes a consolidated probability mask $\boldsymbol{\theta}_e \in [0,1]^s$ from the aggregated data and uses it to generate binary masks $\mathbf{m}_e \in \{0,1\}^s$ through a Bernoulli sampling process (\textcircled{2}-b), which are then sent to the cloud server for further processing (\textcircled{2}-c).

The cloud server plays a pivotal role in integrating the updates received from various edge servers. It performs the final aggregation of the binary masks $\mathbf{m}_e$ uploaded by the edge servers to form a global probability mask $\boldsymbol{\theta}_g \in [0,1]^s$ 
(\textcircled{3}-a). This global mask $\boldsymbol{\theta}_g$ is then broadcast back to all clients through the edge servers (\textcircled{3}-b), facilitating a synchronized update across the network. This mechanism ensures that the global model is continually refined and updated based on the collective learning from all participating clients.

\subsection{Local Training in Client}

In \methodname{}, for round $t$, clients initially receive a global shared probability mask $\boldsymbol{\theta}_{g,t-1}$ from the cloud server. Clients combine $\boldsymbol{\theta}_{g,t-1}$ with $\boldsymbol{\theta}_{k,t-1}^p$ to generate $\boldsymbol{\theta}_{k,t} \in [ 0,1]^d$, which represents the activation probabilities of the model parameters. To optimize within a broader parameter space, this probability mask is transformed into a \textit{score mask} $\mathbf{s} = (s^1,s^2,\ldots,s^d) \in \mathbb{R}^d$ via the inverse Sigmoid function. This transformation allows the \textit{score mask} $\mathbf{s}$ to vary freely within the real number range, providing more flexibility for subsequent optimization. This design permits the \textit{score mask} to flexibly generate the corresponding probability mask by setting $\boldsymbol{\theta} = \text{Sigmoid}(\mathbf{s})$, ensuring each value is appropriately within the [0, 1] interval. The detailed steps for local probability mask training in round $t$ are described as follows and can also be seen in lines 8-21 in Algorithm~\ref{alg:H-FedSN}.

\begin{figure}[htbp]

\centering
\includegraphics[width=0.8\linewidth]{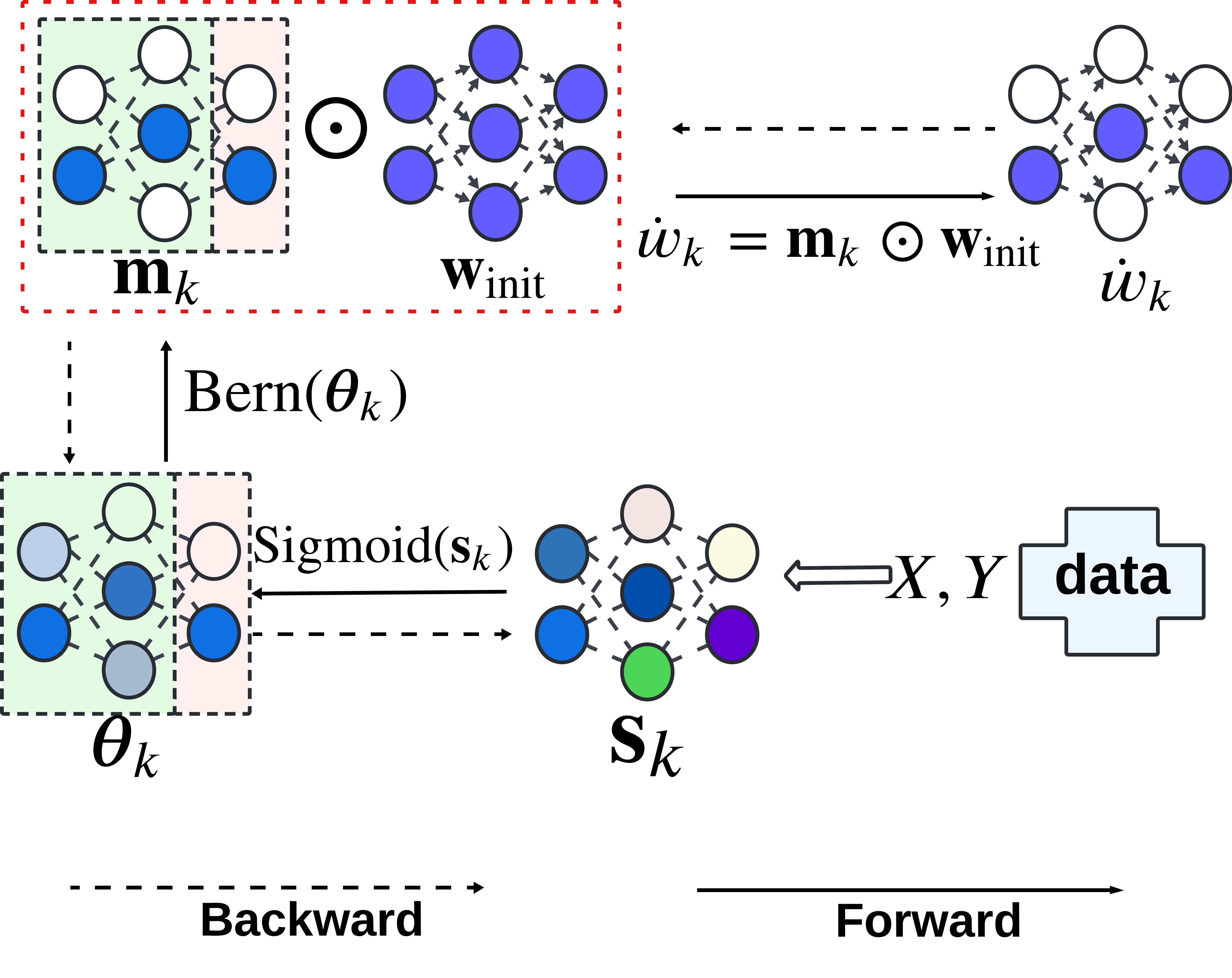} 
\vspace{-0.1in}
\caption{\methodname{} Local Training in Client.}
\label{local_train}
\end{figure}
\begin{enumerate}[leftmargin=*]
    \item Clients receive  the global probability mask $\boldsymbol{\theta}_{g,t-1}$ from the cloud server, then combine $\boldsymbol{\theta}_{g,t-1}$ with $\boldsymbol{\theta}_{k,t-1}^p$ to generate $\boldsymbol{\theta}_{k,t}$. Each client then inputs this probability mask into the inverse Sigmoid function $\text{Sigmoid}^{-1}(\cdot)$, obtaining the \textit{score mask} $\mathbf{s}_{k,t} = \text{Sigmoid}^{-1}(\boldsymbol{\theta}_{k,t})$.
    \item Clients transform back to $\boldsymbol{\theta}_{k,t} = \text{Sigmoid}(\mathbf{s}_{k,t})$, then use Bernoulli sampling to extract binary masks $\mathbf{m}_{k,t}$ from $\boldsymbol{\theta}_{k,t}$.
    \item The sampled binary mask $\mathbf{m}_{k,t}$ is used to sparsify the initial weight vector $\mathbf{w}_{\text{init}}$: $\dot{w}_{k,t} = \mathbf{m}_{k,t} \odot \mathbf{w}_{\text{init}}$.
    \item The $\dot{w}_{k,t}$ is utilized for forward propagation to compute the outputs, and the computed local loss $L(f(\dot{w}_{k,t}), D_k)$ is then used in backpropagation to update the \textit{score mask} according to the formula $\mathbf{s}_{k,t} = \mathbf{s}_{k,t} - \eta \nabla L(f(\dot{w}_{k,t}), D_k)$(where $\eta$ is the local learning rate).
\end{enumerate}
% \vspace{-0.1in}
Fig.~\ref{local_train} illustrates the process of a single epoch of local training on a client. Steps 2 to 4 involve all local operations that are differentiable, except for the Bernoulli sampling step. Since Bernoulli sampling itself is non-differentiable, this would pose a problem in traditional gradient descent frameworks. To address this, we adopt the straight-through estimator technique proposed by \cite{bengio2013estimating}. This technique allows us to "pass-through" gradients during the Bernoulli sampling process: during backpropagation, we set the output gradient of the Bernoulli function to be equal to the input probability mask $\boldsymbol{\theta}_{k,t}$. Thus, although the sampling step is not differentiable, we can still effectively propagate gradients throughout the network and update the score mask $\mathbf{s}_{k,t}$.
After $\tau$ training rounds, each client completes the training of their local probability mask $\boldsymbol{\theta}_{k,t} = \text{Sigmoid}(\mathbf{s}_{k,t})$ and generates a binary mask $\mathbf{m}_{k,t} = \textbf{Bern}(\boldsymbol{\theta}_{k,t})$ through Bernoulli sampling. The client keeps the private part, $\mathbf{m}_{k,t}^p$, and uploads the shared layers, $\mathbf{m}_{k,t}^s$, to the respective connected edge server. The next step involves the edge server aggregating these binary masks to form a comprehensive model update.

\subsection{Bayesian Aggregation in HFL}
In the HFL architecture, both edge and cloud servers implement a Bayesian aggregation strategy to process the binary masks they receive. This approach marks a significant departure from traditional federated learning methods, such as FedAvg, which primarily rely on simple averaging to aggregate updates. Unlike these conventional methods, \methodname{} leverages Bayesian aggregation to more effectively account for the inherent uncertainties in data updates. This capability allows \methodname{} to construct more robust and adaptive global models by synthesizing updates represented as binary masks from clients across multiple layers of the architecture.
The subsequent sections provide a detailed description of the aggregation processes at both the edge server and cloud server levels.
% \begin{figure}[H]
% \centering
% \begin{minipage}[t]{0.49\linewidth}
%     \centering
%     \includegraphics[width=\linewidth]{edge_agg3.png} 
%     \caption{Edge Server Bayesian Aggregation.}
%     \label{edge_agg}
% \end{minipage}
% \hfill 
% \begin{minipage}[t]{0.49\linewidth}
%     \centering
%     \includegraphics[width=\linewidth]{cloud_agg2.png}
%     \caption{Cloud Server Bayesian Aggregation.}
%     \label{cloud_agg}
% \end{minipage}
% \end{figure}
\begin{figure}[H]
\centering
\includegraphics[width=0.7\linewidth]{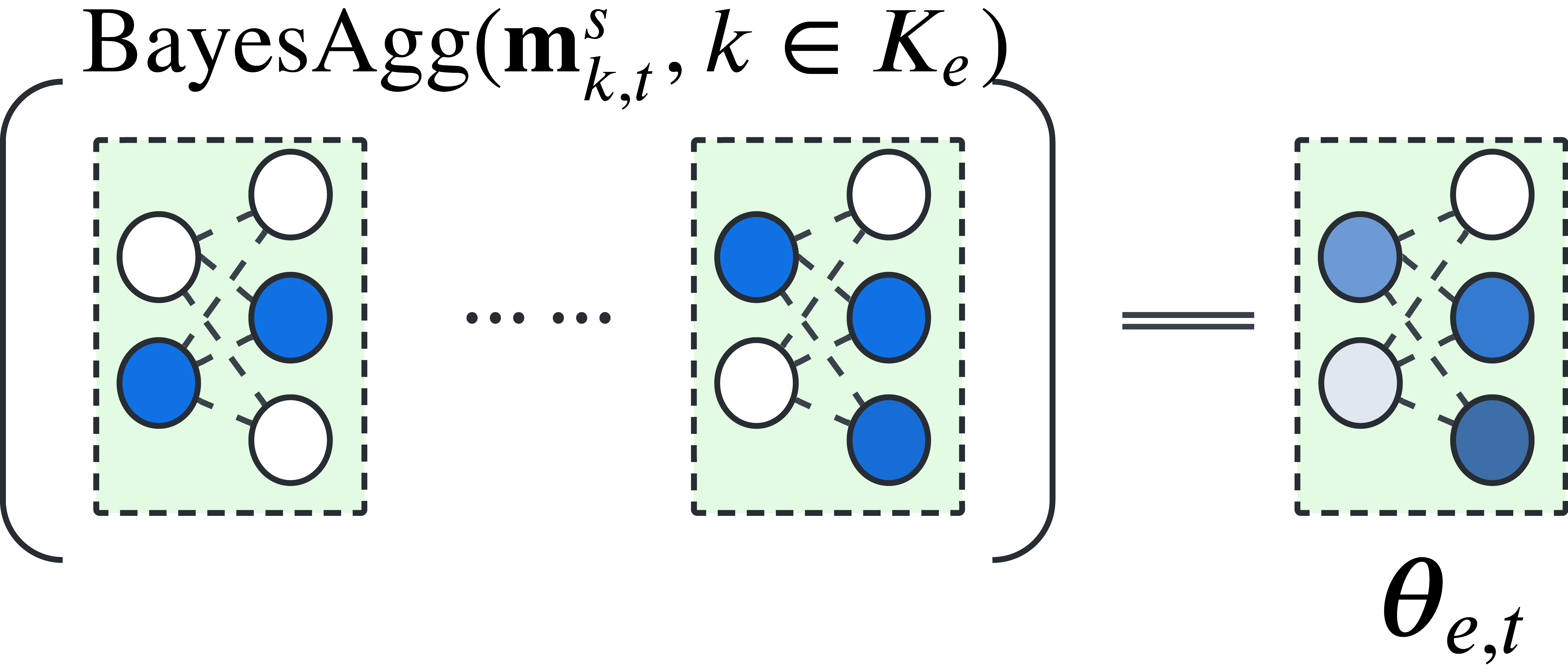}
\caption{Edge Server Bayesian Aggregation.}
\label{edge_agg}
\end{figure}

\subsubsection{Edge Server Bayesian Aggregation}

% As shown in Fig.~\ref{edge_agg}, each edge server $e$, $e \in E$, is responsible for collecting binary masks \(\mathbf{m}_{k,t}^s\) uploaded by all directly connected clients $k$, $k \in K_e$, at round $t$.  The edge server employs a Bayesian aggregation method, as shown in Algorithm~\ref{alg:bayesAggAtEdge}, to integrate these binary masks into an edge probability mask $\boldsymbol{\theta}_{e,t}$. Specifically, the probability mask $\boldsymbol{\theta}_{e,t}$ is modeled using the parameters of a Beta distribution, $\alpha_{e,t}$ and $\beta_{e,t}$, which are initially set to $\alpha_{e,0} = \beta_{e,0} = \lambda_0$.
Fig.~\ref{edge_agg} shows that each edge server $e \in E$ collects binary masks $\mathbf{m}_{k,t}^s$ from its directly connected clients $k \in K_e$ at round $t$ and aggregates them into an edge-level probability mask $\boldsymbol{\theta}_{e,t}$ using Bayesian aggregation (Algorithm\ref{alg:bayesAggAtEdge}). Specifically, $\boldsymbol{\theta}_{e,t}$ is modeled by a Beta distribution with parameters $\alpha_{e,t}$ and $\beta_{e,t}$, both initialized as $\lambda_0 = 1$ at round $t=0$.

At round $t=0$, the system has no prior knowledge, the probability mask follows a uniform distribution over [0, 1]. Upon receiving client masks $\mathbf{m}_{k,t}^s$ from client $k \in K_e$, edge $e$ updates the prior to a posterior. Exploiting the conjugacy of Beta and Bernoulli distributions, the posterior remains Beta-distributed with updated parameters:
Since Beta and Bernoulli distributions form a conjugate pair, the posterior conveniently remains a Beta distribution with updated parameters: $\alpha_{e,t} = \alpha_{e,t-1} + \sum_{k \in K_e} \mathbf{m}_{k,t}^s$  and $\beta_{e,t} = \beta_{e,t-1} + |K_e| \cdot \mathbf{1} - \sum_{k \in K_e} \mathbf{m}_{k,t}^s$,
where $\mathbf{m}_{k,t}^s$ represents binary mask submitted by the $k$-th client under edge server $e$ during round $t$. The $\mathbf{1}$ is the \textit{s}-dimensional all-ones vector. $|K_e|$ denotes the number of clients connected to edge server $e$. Subsequently, the edge server calculates the probability mask $\boldsymbol{\theta}_{e,t}$ based on the updated parameters $\alpha_{e,t}$ and $\beta_{e,t}$ :
$
\boldsymbol{\theta}_{e,t} = \frac{\alpha_{e,t}- 1}{\alpha_{e,t} + \beta_{e,t} - 2}
$
. This operation is applied element-wise. The edge server then broadcasts $\mathbf{m}_{e,t} = \text{Bern}(\boldsymbol{\theta}_{e,t})$  to the cloud. 
To avoid overfitting the model to historical data, $\alpha_{e,t}$ and $\beta_{e,t}$ are reset to their initial value $\lambda_0 = 1$ at the start of each $10^{th}$ training round. This historical window size $w$ provides a practical frequency for incorporating recent updates while preventing older statistics from dominating indefinitely. We provide an ablation on $w$ in Appx.~\ref{sec:supp_window_ablation}.
For the pseudocode of edge server Bayesian aggregation, see Algorithm~\ref{alg:bayesAggAtEdge}. 
% \begin{algorithm}[h!]
% \begin{algorithmic}[1]
% \REQUIRE Binary masks $\mathbf{m}_{k,t}^s$ from clients $k \in K_e$, and round number $t$
% \ENSURE Updated binary mask $\boldsymbol{m}_{e,t}$
% \IF {ResPriors(t)}
%         \STATE $\alpha_{e,t-1} = \beta_{e,t-1} = \lambda_0$
% \ENDIF
% \STATE $\alpha_{e,t} = \alpha_{e,t-1} + \sum_{k \in K_e} \mathbf{m}_{k,t}^s$
% \STATE $\beta_{e,t} = \beta_{e,t-1} + |K_e| \cdot \mathbf{1} - \sum_{k \in K_e} \mathbf{m}_{k,t}^s$
% \STATE $\boldsymbol{\theta}_{e,t} = \frac{\alpha_{e,t} - 1}{\alpha_{e,t} + \beta_{e,t} - 2}$
% \STATE $\boldsymbol{m}_{e,t} \leftarrow 
%   \textbf{Bern}(\boldsymbol{\theta}_{e,t})$
% \STATE \textbf{Output:}  $\boldsymbol{m}_{e,t} $
% \caption{Bayesian Aggregation At Edge}
% \label{alg:bayesAggAtEdge}
% \end{algorithmic}
% \end{algorithm}
\begin{algorithm}[h!]
\small
\caption{Bayesian Aggregation at Edge}
\label{alg:bayesAggAtEdge}
\begin{algorithmic}[1]
\REQUIRE Binary masks $\mathbf{m}_{k,t}^s$ from clients $k \in K_e$, round number $t$, reset interval $w$
\ENSURE Updated binary mask $\boldsymbol{m}_{e,t}$
\IF {$t = 0$ \textbf{or} $t \bmod w = 0$}
    \STATE $\alpha_{e,t-1} = \beta_{e,t-1} = \lambda_0$ \COMMENT{reset Beta priors}
\ENDIF
\STATE $\alpha_{e,t} = \alpha_{e,t-1} + \sum_{k \in K_e} \mathbf{m}_{k,t}^s$
\STATE $\beta_{e,t} = \beta_{e,t-1} + |K_e| \cdot \mathbf{1} - \sum_{k \in K_e} \mathbf{m}_{k,t}^s$
\STATE $\boldsymbol{\theta}_{e,t} = \frac{\alpha_{e,t} - 1}{\alpha_{e,t} + \beta_{e,t} - 2}$
\STATE $\boldsymbol{m}_{e,t} \leftarrow \textbf{Bern}(\boldsymbol{\theta}_{e,t})$
\STATE \textbf{Output:} $\boldsymbol{m}_{e,t}$
\end{algorithmic}
\end{algorithm}

\subsubsection{Cloud Server Bayesian Aggregation}

The cloud server is responsible for collecting the binary masks uploaded by all edge servers $e$, $e \in E$, and performs a global aggregation. 

\begin{figure}[H]
\centering
\includegraphics[width=0.7\linewidth]{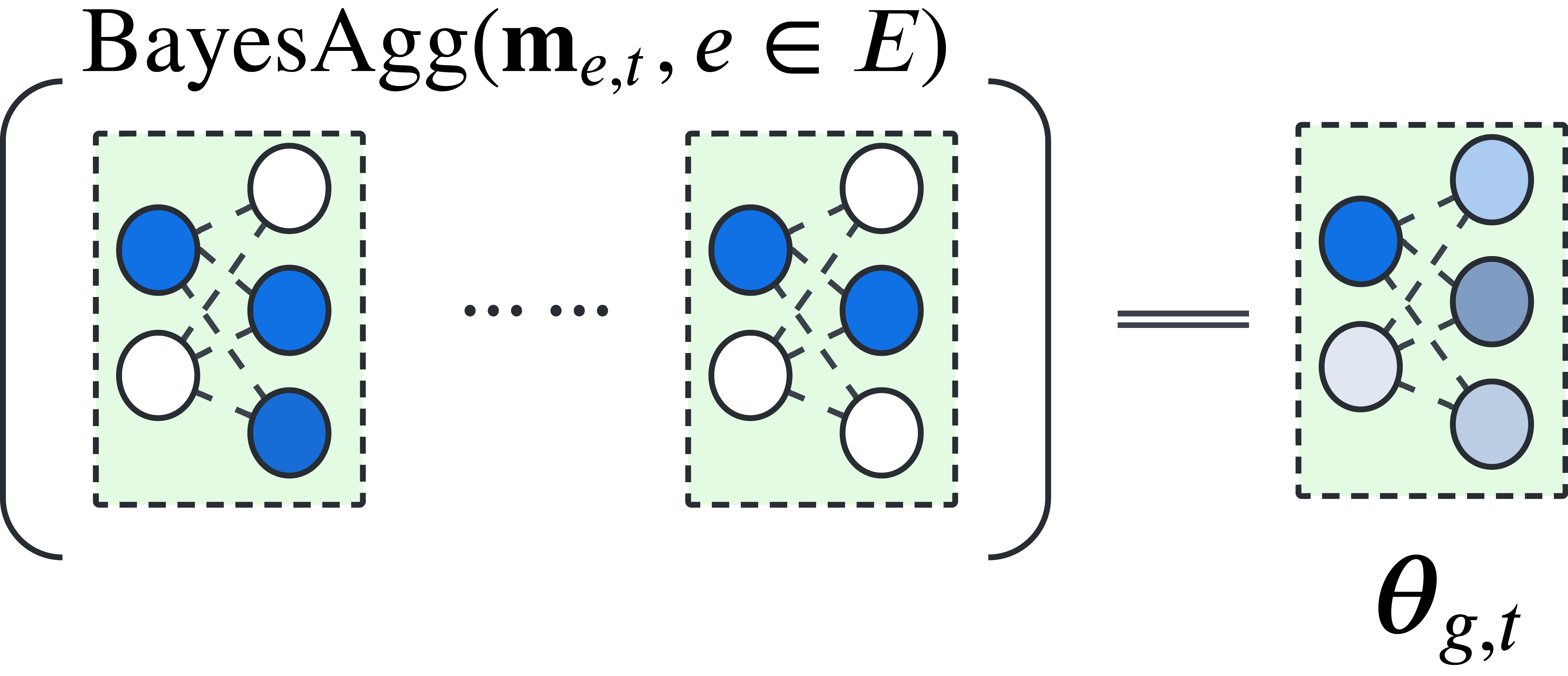}
\caption{Cloud Server Bayesian Aggregation.}
\label{cloud_agg}
\end{figure}

At the cloud level, as illustrated in Fig.~\ref{cloud_agg} and Algorithm~\ref{alg:bayesAggAtCloud}, the global probability mask $\boldsymbol{\theta}_{g,t}$ is modeled using a Beta distribution $\text{Beta}(\alpha_{g,t}, \beta_{g,t})$, with parameters $\alpha_{g,t}$ and $\beta_{g,t}$ related to the training round and initially set to $\alpha_{g,0} = \beta_{g,0} = \lambda_0$. The parameters for the updated posterior distribution are calculated as follows:
$\alpha_{g,t} = \alpha_{g,t-1} + \sum_{e \in E} \mathbf{m}_{e,t}$ and $\beta_{g,t} = \beta_{g,t-1} + |E| \cdot \mathbf{1} - \sum_{e \in E} \mathbf{m}_{e,t}$
% \begin{align*}
% \alpha_{g,t} = \alpha_{g,t-1} + \sum_{e \in E} \mathbf{m}_{e,t} \\
% \beta_{g,t} = \beta_{g,t-1} + |E| \cdot \mathbf{1} - \sum_{e \in E} \mathbf{m}_{e,t}
% \end{align*}
, where $\mathbf{m}_{e,t}$ represents the binary mask submitted by the $e$-th edge server to the cloud during round $t$. 
The $\mathbf{1}$ is the \textit{s}-dimensional all-ones vector. $|E|$ denotes the number of edge servers. Following the parameter update, the cloud server computes the mode of the Bernoulli distributions to derive the global probability mask $\boldsymbol{\theta}_{g,t}$, as suggested by \cite{ferreira2021bayesian}:
$
\boldsymbol{\theta}_{g,t} = \frac{\alpha_{g,t} - 1}{\alpha_{g,t} + \beta_{g,t} - 2}
$
. 
Subsequently, the cloud broadcasts the probability mask $\boldsymbol{\theta}_{g,t}$ to all clients through the edge servers. To avoid overfitting the model to historical data, $\alpha_{g,t}$ and $\beta_{g,t}$ are reset to their initial value $\lambda_0 = 1$ at the start of each $10^{th}$ training round, which provides a practical frequency for incorporating recent updates while preventing older statistics from dominating indefinitely. For the pseudocode of cloud server Bayesian aggregation, see Algorithm~\ref{alg:bayesAggAtCloud}.
% \begin{algorithm}[H]
% \begin{algorithmic}[1]
% \REQUIRE Binary masks $\mathbf{m}_{e,t}$ from edge servers $e \in E$, and round number $t$
% \ENSURE Updated global probability mask $\boldsymbol{\theta}_{g,t}$
% \IF {ResPriors(t)}
%         \STATE $\alpha_{e,t-1} = \beta_{e,t-1} = \lambda_0$
% \ENDIF
% \STATE $\alpha_{g,t} = \alpha_{g,t-1} + \sum_{e \in E} \mathbf{m}_{e,t}$
% \STATE $\beta_{g,t} = \beta_{g,t-1} + |E| \cdot \mathbf{1} - \sum_{e \in E} \mathbf{m}_{e,t}$
% \STATE $\boldsymbol{\theta}_{g,t} = \frac{\alpha_{g,t} - 1}{\alpha_{g,t} + \beta_{g,t} - 2}$
% \STATE \textbf{Output:} $\boldsymbol{\theta}_{g,t}$
% \caption{Bayesian Aggregation At Cloud}
% \label{alg:bayesAggAtCloud}
% \end{algorithmic}
% \end{algorithm}
\begin{algorithm}[H]
\small
\caption{Bayesian Aggregation at Cloud}
\label{alg:bayesAggAtCloud}
\begin{algorithmic}[1]
\REQUIRE Binary masks $\mathbf{m}_{e,t}$ from edge servers $e \in E$, round number $t$, reset interval $w$
\ENSURE Updated global probability mask $\boldsymbol{\theta}_{g,t}$
\IF {$t = 0$ \textbf{or} $t \bmod w = 0$}
    \STATE $\alpha_{g,t-1} = \beta_{g,t-1} = \lambda_0$ \COMMENT{reset Beta priors}
\ENDIF
\STATE $\alpha_{g,t} = \alpha_{g,t-1} + \sum_{e \in E} \mathbf{m}_{e,t}$
\STATE $\beta_{g,t} = \beta_{g,t-1} + |E| \cdot \mathbf{1} - \sum_{e \in E} \mathbf{m}_{e,t}$
\STATE $\boldsymbol{\theta}_{g,t} = \frac{\alpha_{g,t} - 1}{\alpha_{g,t} + \beta_{g,t} - 2}$
\STATE \textbf{Output:} $\boldsymbol{\theta}_{g,t}$
\end{algorithmic}
\end{algorithm}

\section{EXPERIMENTS}

We empirically evaluate the performance of \methodname{}, focusing on both accuracy and communication costs. Our experimental analysis covers four distinct datasets: the widely-utilized MNIST \cite{lecun1998gradient}, supplemented by three IoT-specific datasets: WISDM (watch) \cite{weiss2019smartphone}, WIDAR \cite{widardata2020,zheng2019zero}, and WISDM (phone) \cite{lockhart2011design}. MNIST acts as a general benchmark, while the inclusion of IoT datasets showcases our method’s applicability in real-world IoT settings. 

We compare \methodname{} against 6 baselines adapted for the HFL architecture.  ESPerHFL~\cite{Ma_2024}, HierFAVG~\cite{liu2020client} naturally supports HFL, whereas FedPer \cite{arivazhagan2019federated}, FedRS \cite{10.1145/3447548.3467254}, FedCAMS \cite{Wang2022Communication-Efficient}, and TOPK \cite{aji2017sparse}, initially designed for standard FL, have been modified to fit an HFL architecture.

\subsection{Datasets Setting}
\subsubsection{Datasets and Preprocessing}

Table~\ref{data_stat} shows the statistical distribution of the datasets used in our experiments.
Below, we describe each dataset along with its preprocessing methods:
\begin{table}[htbp]
    \centering
    \vspace{-0.1in}
    \caption{Statistical information of datasets.}
    \vspace{-0.1in}
    \scalebox{0.65}{
    \setlength{\tabcolsep}{3pt} 
    \renewcommand{\arraystretch}{1.0} 
    \begin{tabular}{@{}llllr@{}} 
        \toprule
        \textbf{Dataset} & \textbf{Dimension} & \textbf{\# Samples (Train / Test)} & \textbf{Classes}  & \#Fixed Classes per Client \\ 
        \midrule
        MNIST & 28 x 28 & 500,000  /  10,000 & 10  & 6\\
        WISDM (WATCH) & 200 x 6 & 16,569  /  4,103 & 12 & 7 \\
        WIDAR & 22 x 20 x 20 & 11,372  /  5,222 & 9  & 5 \\
        WISDM (PHONE) & 200 x 6 & 13,714  /  4,073 & 12 & 7\\
        \bottomrule
    \end{tabular}}
    \label{data_stat}
    \vspace{-0.1in}
\end{table}
\begin{itemize}[leftmargin=*]
\item \textbf{MNIST} comprises handwritten digits across 10 classes, widely used for image classification model training \cite{lecun1998gradient}. Preprocessing involved converting images to tensors and normalizing pixel values to [-1, 1].
    
\item \textbf{WISDM (WATCH)} \cite{weiss2019smartphone} encompasses accelerometer and gyroscope readings from smartwatches, capturing 18 daily activities performed by 51 participants. Our preprocessing approach aligned with established methods in the field of sensor-based activity recognition \cite{ravi2005activity, reyes2016transition, ronao2016human}. We segmented each 3-minute recording using a sliding window technique, employing a 10-second window with 50\% overlap. Subsequently, we applied normalization to each dimension of the resulting samples, centering the data around zero and adjusting the scale to achieve unit variance.

\item \textbf{WIDAR} \cite{widardata2020,zheng2019zero}, designed for contactless gesture recognition using Wi-Fi signal strength, includes data from 17 participants performing 22 unique gestures. We applied body velocity profiling (BVP) to account for environmental variations, followed by scalar normalization. The final representation was structured into 22 × 20 × 20 samples, covering the time axis and x-y velocity features. Only gestures recorded by more than three users were retained to ensure consistency between training and testing sets.

\item \textbf{WISDM (PHONE)} \cite{lockhart2011design} consists of accelerometer data obtained from smartphones during participants' daily activities. We implemented an identical preprocessing protocol to that used for the WISDM (WATCH) dataset. This involved segmenting the data with a 10-second sliding window (50\% overlap), followed by normalization of each dimension to zero mean and unit variance.

\end{itemize}

\subsubsection{Data Partitioning and Non-IID Setting}

\begin{figure}[htbp]
\centering
\includegraphics[width=0.9\linewidth]{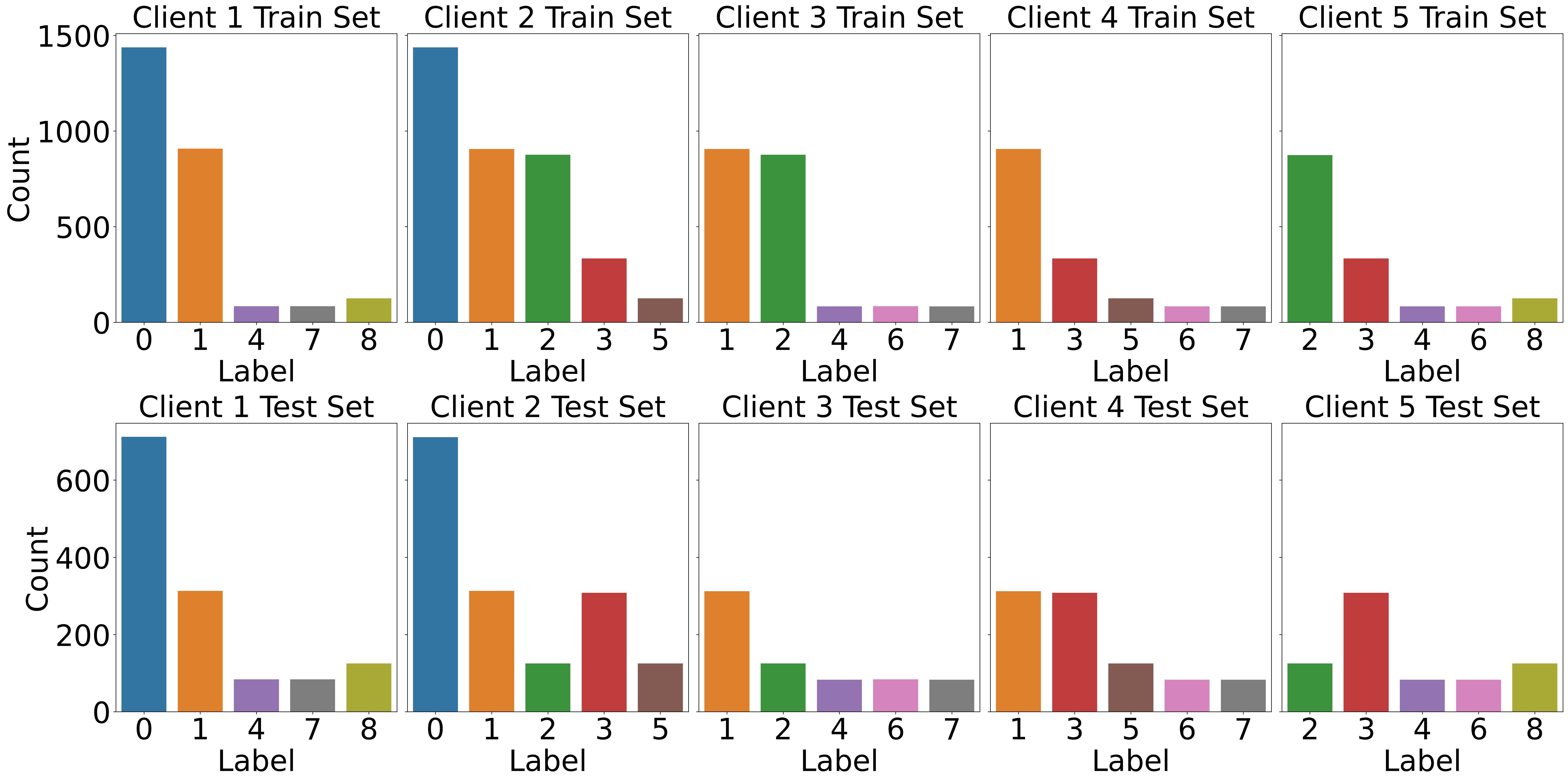}  
\caption{WIDAR Dataset NoN-IID Data Distribution Across 5 Clients}
\label{fig:non-iid_dis}
\end{figure}

We simulate non-IID data distributions using a label imbalance approach that ensures each client has a fixed set of labels, following FedAvg \cite{mcmahan2017communication} and further refined by Li et al.\cite{li2022federated}. Specifically, each client is randomly assigned n distinct label IDs, and the samples of each label are distributed equally among the clients possessing that label. This guarantees every client consistently holds data from exactly n labels with no overlap among clients. We apply the same label partitioning to both training and test sets to maintain consistent levels of label imbalance. Table~\ref{data_stat} shows the total number of classes for each dataset and the fixed number of classes per client. For illustration, Fig.~\ref{fig:non-iid_dis} depicts how WIDAR is partitioned among 5 clients.

\subsection{Experimental Design}
\subsubsection{\methodname{} and Baselines Setting}
\ 
\textbf{\methodname{}}. We used a 4-layer convolutional neural network (CNN) comprising two convolutional blocks (each with two convolutional layers followed by a pooling layer, with 64 filters per layer in the first block and 128 in the second) and three fully-connected layers (two with 256 units each, followed by an output layer matching the dataset's number of classes). For \methodname{}, we set the last three layers as private layers, which do not participate in the aggregation process.

To comprehensively evaluate the performance of \methodname{}, we compare \methodname{} against 6 baselines:
\begin{itemize}[leftmargin=*]
    \item \textbf{ESPerHFL}~\cite{Ma_2024}: Implements edge-server personalization in HFL by using learnable mixing parameters and similarity-based aggregation with Tanimoto coefficients.
    
\item \textbf{HierFAVG} \cite{liu2020client} Hierarchical federated averaging with client-edge-cloud structure.
\item \textbf{FedPer} \cite{arivazhagan2019federated} Federated personalization keeping part of model parameters local.
\item \textbf{FedRS} \cite{10.1145/3447548.3467254} Addresses label distribution skew using Restricted Softmax.
\item \textbf{FedCAMS} \cite{Wang2022Communication-Efficient} Combines gradient compression with adaptive optimization to reduce communication overhead.
\item \textbf{TOPK} \cite{aji2017sparse} Sparsification method selecting largest k (3.125\%) elements of gradients.
\end{itemize}
All baselines are implemented with aggregations at both edge and cloud levels in the HFL architecture.

In addition, we adopted the same base model configuration for each baseline method, which is the same 4-layer CNN (CONV-4) used for \methodname{}. We also applied the same data configurations, ensuring consistency across all experiments with the non-IID data distribution.

\subsubsection{Performance Metrics}
\label{sec:perfomance_metrics}
To comprehensively assess the efficacy of our proposed algorithm, we evaluate the following key performance metrics:

\begin{itemize}[leftmargin=*]
  \item \textbf{Inference Performance:} We measure inference performance in two ways: (1) \textbf{Average Accuracy}: Defined as the mean accuracy across all clients, this metric quantifies the overall inference performance of the federated model. It serves as the primary benchmark for comparing methods, clearly indicating the algorithm's effectiveness in achieving accurate predictions on distributed, heterogeneous datasets.
  (2) \textbf{Accuracy Distribution across Clients}:  In addition to the average accuracy, we visualize the distribution of inference accuracies among individual clients using boxplots. This metric captures the consistency and robustness of the algorithm across heterogeneous and imbalanced client distributions, reflecting its practical reliability in realistic IoT environments.

  \item \textbf{Communication Cost:} We measure the volume of data (in KB) transmitted per training round between clients and edge servers, and from edge servers to the cloud. Minimizing communication overhead is critical in real-world IoT scenarios due to bandwidth limitations, directly affecting system efficiency, scalability, and latency.
\end{itemize}

\subsubsection{Experimental Setting}
\label{exp:exp_setting}
To illustrate the adaptability of our proposed algorithm across different scenarios and environments, we designed the following experimental configurations:

\begin{itemize}[leftmargin=*]
    \item \textbf{E2C5:} 2 edge nodes and 5 client nodes. This minimalistic setup tests the algorithm's effectiveness in small-scale deployments, such as some small offices or local retail stores with limited devices and minimal infrastructure~\cite{purohit2021evaluation}.
    \item \textbf{E20C50:} 20 edge nodes and 50 client nodes. This configuration evaluates scalability and robustness in larger, complex network topologies, simulating environments like a smart city surveillance system across multiple buildings~\cite{anthopoulos2015understanding}.
    \item \textbf{E2C50:} 2 edge nodes and 50 client nodes. This setting simulates high client density scenarios, such as a smart healthcare system for remote patient monitoring where numerous wearable devices and home health monitors connect to local healthcare facility servers~\cite{8066704}.
    \item \textbf{Imbalanced E5C50:} 5 edge nodes with 50 clients distributed in ratios of 0.4, 0.2, 0.2, 0.1, and 0.1. This imbalanced setup tests consistency and fairness across unevenly distributed network nodes, representing scenarios like smart agriculture systems with varying device densities across different farm sections~\cite{Virk2020Smart}.
\end{itemize}

%highlight
For all experiments, the hyperparameters were consistently set as follows: Global Rounds = 200, the learning rates for \methodname{} were 0.01 for all datasets except the WIDAR dataset, where the learning rate was 0.035. For the baseline algorithms, the learning rate was uniformly set to 0.001 (see Appx.~\ref{appx:lr}).
%highlight

\begin{table*}[t!]
\setlength{\abovecaptionskip}{3pt}
\caption{Performance comparison of H-FedSN and baseline methods across four datasets in E2C5 configuration. Methods are evaluated under two metrics as described in \S\ref{sec:perfomance_metrics} with avg Acc. (average accuracy of clients; higher (\raisebox{0.2ex}{$\uparrow$}), the better) and Comm. (communication cost; lower (\raisebox{0.2ex}{$\downarrow$}), the better) as the metrics. The \colorbox{gray!20}{\textbf{Best}} results are highlighted in bold, the \underline{\underline{{second-best}}} results are double-underlined, and the \underline{third-best} results are single-underlined.}

\label{tab:e2_c5_comm_and_acc}
\centering
\setlength{\tabcolsep}{7pt}
\renewcommand{\arraystretch}{1}
\fontsize{8}{12}\selectfont 
\begin{tabular}{c cc cc cc cc}
\toprule
\multicolumn{1}{c}{\multirow{2}{*}{\textbf{Algorithms}}} & \multicolumn{2}{c}{\raisebox{0.3ex}{\textbf{MNIST}}} & \multicolumn{2}{c}{\raisebox{0.3ex}{\textbf{WISDM(WATCH)}}} & \multicolumn{2}{c}{\raisebox{0.3ex}{\textbf{WIDAR}}} & \multicolumn{2}{c}{\raisebox{0.3ex}{\textbf{WISDM(PHONE)}}}  \\ \cline{2-9}
\multicolumn{1}{c}{} & \raisebox{-0.3ex}{avg Acc.} \raisebox{-0.1ex}{$\uparrow$} & \raisebox{-0.3ex}{Comm. (KB)} \raisebox{-0.1ex}{$\downarrow$} & \raisebox{-0.3ex}{avg Acc.} \raisebox{-0.1ex}{$\uparrow$} & \raisebox{-0.3ex}{Comm. (KB)} \raisebox{-0.1ex}{$\downarrow$} & \raisebox{-0.3ex}{avg Acc.} \raisebox{-0.1ex}{$\uparrow$} & \raisebox{-0.3ex}{Comm. (KB)} \raisebox{-0.1ex}{$\downarrow$} & \raisebox{-0.3ex}{avg Acc.} \raisebox{-0.1ex}{$\uparrow$} & \raisebox{-0.3ex}{Comm. (KB)} \raisebox{-0.1ex}{$\downarrow$} \\ \midrule
\multicolumn{1}{l}{ESPerHFL} & 0.9963 & 120828.63 & 0.7005 & 14168.75 & 0.7532 & 72416.56 & 0.3882 & 14168.75 \\ 
\multicolumn{1}{l}{HierFAVG} & 0.9968 & 60414.31 & \underline{{0.7382}} & 7084.38 & \underline{\underline{{0.7789}}} & 36208.28 & 0.3707 & 7084.38\\
\midrule
\multicolumn{1}{c}{} & \multicolumn{8}{c}{\raisebox{0.3ex}{\textbf{Personalization-enhanced FL methods}}} \\ 
\multicolumn{1}{l}{FedPer} & 0.9950 & 60334.00 & 0.7246 & 6988.00 & \colorbox{gray!20}{\textbf{0.8285}} & 36136.00 &\colorbox{gray!20}{\textbf{0.4836}} & 6988.00\\
\multicolumn{1}{l}{FedRS} & 0.9965 & 60414.31 & \colorbox{gray!20}{\textbf{0.7597}} & 7084.38 & 0.7432 & 36208.28 & 0.3814 & 7084.38\\ \midrule
\multicolumn{1}{c}{} & \multicolumn{8}{c}{\raisebox{0.3ex}{\textbf{ Communication-optimizated FL methods}}} \\ 
\multicolumn{1}{l}{FedCAMS} & 0.9929 & \underline{1887.98} & 0.5965 & \underline{211.42} & 0.7097 & \underline{1131.54} & \underline{{0.3967}} & \underline{211.42}\\
\multicolumn{1}{l}{TOPK} & 0.9908 & \underline{\underline{{1887.98}}} & 0.5977 & \underline{\underline{{211.41}}} & 0.6218 & \underline{\underline{{1131.53}}}  & 0.3002 & \underline{\underline{{211.41}}} \\
\midrule
\multicolumn{1}{l}{\methodname{}(\textbf{Our})} & 0.9936 & \colorbox{gray!20}{\textbf{252.94}}& \underline{\underline{{0.7442}}} & \colorbox{gray!20}{\textbf{121.88}} & \underline{{0.7637}} & \colorbox{gray!20}{\textbf{264.75}} & \underline{\underline{{0.4298}}} & \colorbox{gray!20}{\textbf{121.88}} \\

\bottomrule
\end{tabular}
\end{table*}

\subsection{Results and Analysis}
In this section, we present and analyze the results obtained from our experiments. We provide a comparative evaluation of \methodname{} against the baselines in terms of inference performance and communication costs.
\subsubsection{E2C5 Configuration}

\begin{figure*}[h!]
    \centering
    \begin{minipage}[t]{0.48\textwidth}
        \centering
        \includegraphics[width=\linewidth]{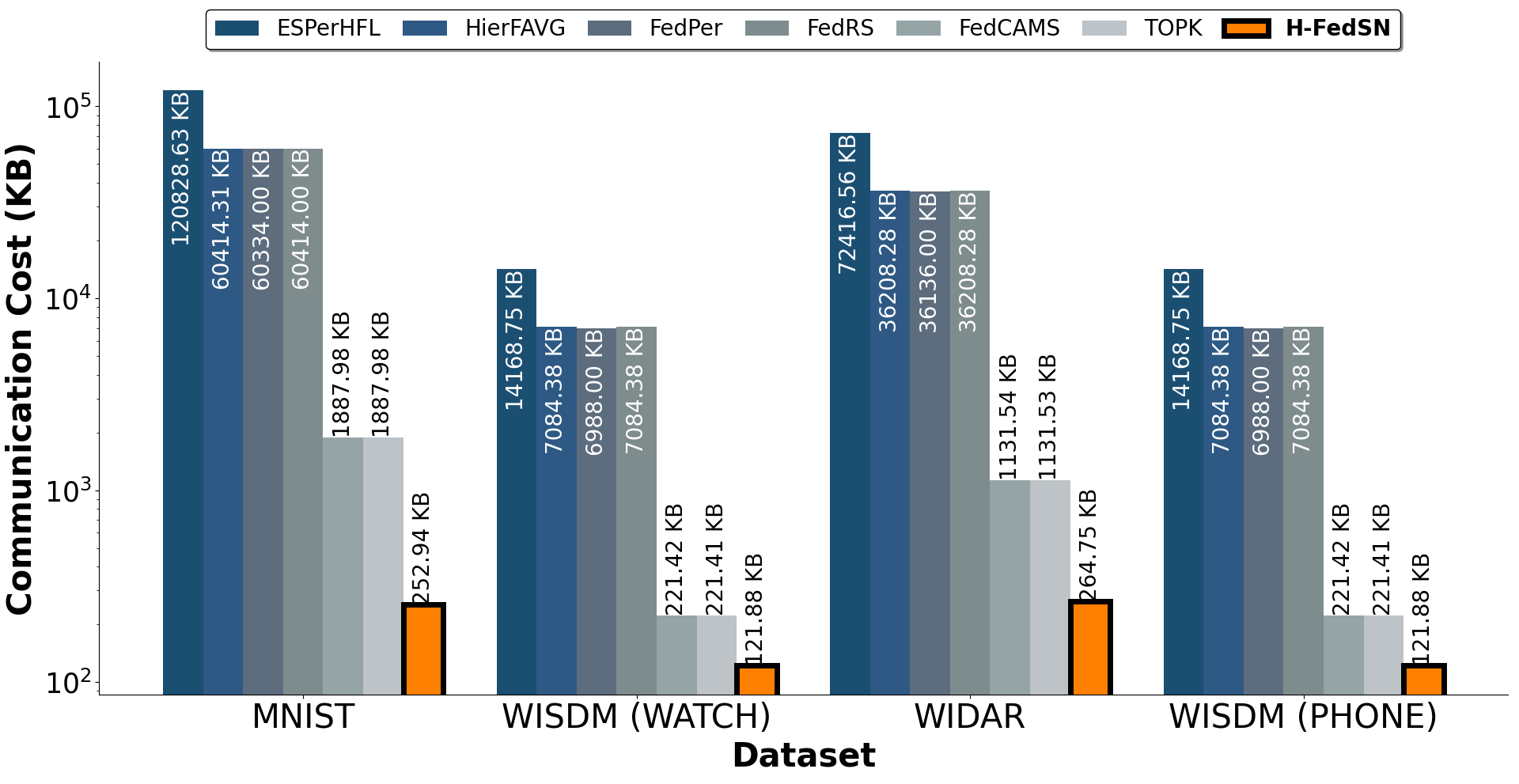}
        \caption{Communication cost for different algorithms across different datasets in all settings.}
        \label{fig:comm}
    \end{minipage}
    \hfill
    \begin{minipage}[t]{0.5\textwidth}
        \centering
        \includegraphics[width=\linewidth]{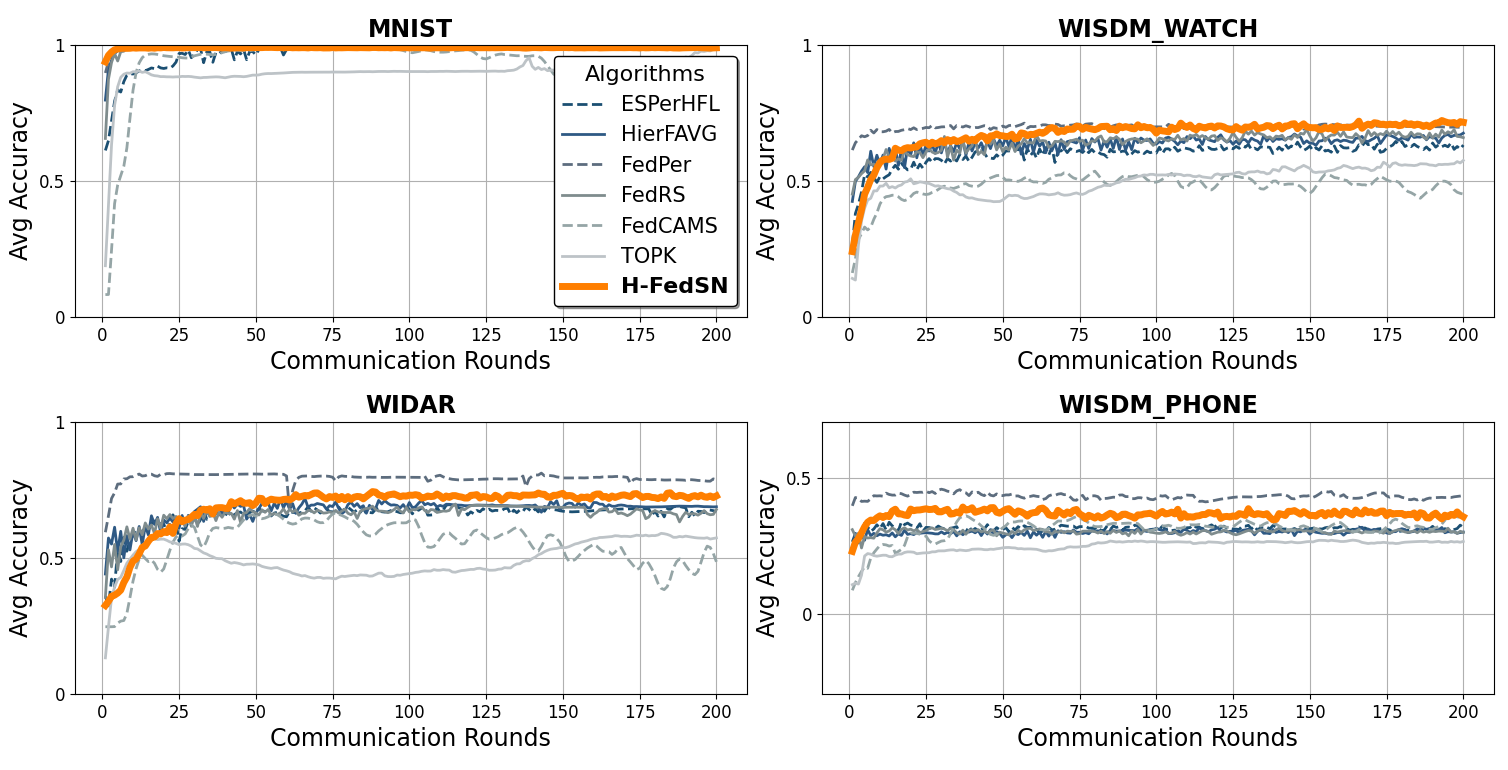}
        \caption{Average accuracy of all clients per round under the \textbf{E2C5} setting across four datasets.}
        \label{fig:client_accuracy_all_datasets}
    \end{minipage}
\end{figure*}
In the \textbf{E2C5} experimental setting, we conduct a comprehensive comparison across three categories of methods: personalization-enhanced approaches (FedPer, FedRS), communication optimization approaches (FedCAMS, TOPK), and the standard hierarchical federated averaging baseline (HierFAVG). As shown in Table.~\ref{tab:e2_c5_comm_and_acc}, we evaluate each method in terms of the average accuracy (avg Acc.) across all clients (higher is better) and the per-round communication cost (Comm.) measured in KB (lower is better). Among all baselines, the personalization-enhanced methods typically achieve high accuracies, so we compare \methodname{} against each dataset’s strongest personalization baseline to gauge how closely it can match their accuracy at significantly lower overhead. For WISDM(WATCH) dataset, FedRS attains an accuracy of 0.7597 (the highest among baselines), whereas \methodname{} reaches 0.7442, only slightly lower, yet FedRS’s communication cost (7084.38 KB) is 58 times larger than \methodname{}’s 121.88 KB. A simliar pattern also appears in WIDAR and WISDM(PHONE) dataset. 
In WIDAR, FedPer's accuracy surpassing \methodname{} a little, but it transmits 36136 KB compared to \methodname{}’s 264.75 KB, over a 136-fold difference. In WISDM(PHONE), where FedPer exceeds \methodname{} a little, yet requires more than 57 times the communication cost (6988 KB vs. 121.88 KB).
Finally, although all baselines achieve relatively similar average accuracy on \textbf{MNIST}, \methodname{} remains below 300KB in per-round communication. In all these comparisons, \methodname{} delivers near-top accuracy while reducing communication by one to two orders of magnitude, thanks to transmitting compact binary masks for shared layers, retaining personalized parameters entirely local, and employing Bayesian aggregation for robust updates.

In Figure~\ref{fig:client_accuracy_all_datasets}, we illustrate the training of six methods (HierFAVG, FedPer, FedRS, FedCAMS, TOPK, and \methodname{}) in HFL architecture. From the plotted curves, it is evident that all methods reach near-convergence around the 20th round, with only minor accuracy fluctuations thereafter, indicating that their global models stabilize at this stage. Notably, TOPK and FedCAMS exhibit non-monotonic behavior in their accuracy curves. TOPK’s accuracy rises rapidly in the early rounds but then undergoes a marked drop before gradually recovering. This pattern emerges because TOPK transmits only the largest gradient components, initially boosting accuracy but overlooking smaller yet important updates. As these neglected updates are later transmitted through error compensation, the model corrects its early bias, causing a dip and eventual rebound. By contrast, FedCAMS combines adaptive optimization with gradient compression, resulting in more moderate updates during the initial phase. Nevertheless, as training proceeds, conflicting gradients arising from multi-layer data heterogeneity lead to a mid-round accuracy dip, which is then partially offset by FedCAMS’s adaptive feedback mechanism. Despite these transient fluctuations, both methods ultimately achieve convergence on par with the other approaches in our experiments.

\noindent
% Next, we now investigate more demanding network conditions featuring larger scales, uneven client distributions, and more complex connectivity structures. In these subsequent experiments (e.g. E2C50, Imbalanced E5C50, E20C50), we narrow our baseline set to FedCAMS and TOPK, two methods specifically designed to reduce communication overhead. As shown in the following sections, H-FedSN not only delivers lower per-round transmission costs than FedCAMS and TOPK, but also achieves higher average accuracy and more consistent performance across heterogeneous clients under diverse large-scale and imbalanced network configurations.
%highlight
We next examine more demanding network conditions with larger scales, uneven client distributions, and more complex connectivity. The subsequent experiments (E2C50, Imbalanced E5C50, E20C50) compare \methodname{} against four representative baselines spanning two categories: FedPer and FedRS for personalization, and FedCAMS and TOPK for communication efficiency. As shown in the following sections, \methodname{} not only delivers lower per-round transmission cost than FedCAMS and TOPK along with higher average accuracy and more consistent per-client performance, but also matches the accuracy of FedPer and FedRS at roughly two orders of magnitude lower transmission cost. 
%highlight

\subsubsection{Imbalanced E5C50 Configuration}

In the \textbf{Imbalanced E5C50} configuration, where client distribution is uneven, \methodname{} demonstrates its robustness by maintaining high performance across varying client densities while significantly reducing communication costs across all datasets. 
\begin{figure}[h!]
    \centering
    % \subfloat[MNIST]{{\includegraphics[width=0.5\linewidth]{exp_res/im_e5_c50/e5-c50-MNIST.png} }}
    \subfloat[WISDM(WATCH)]{{\includegraphics[width=0.33\linewidth]{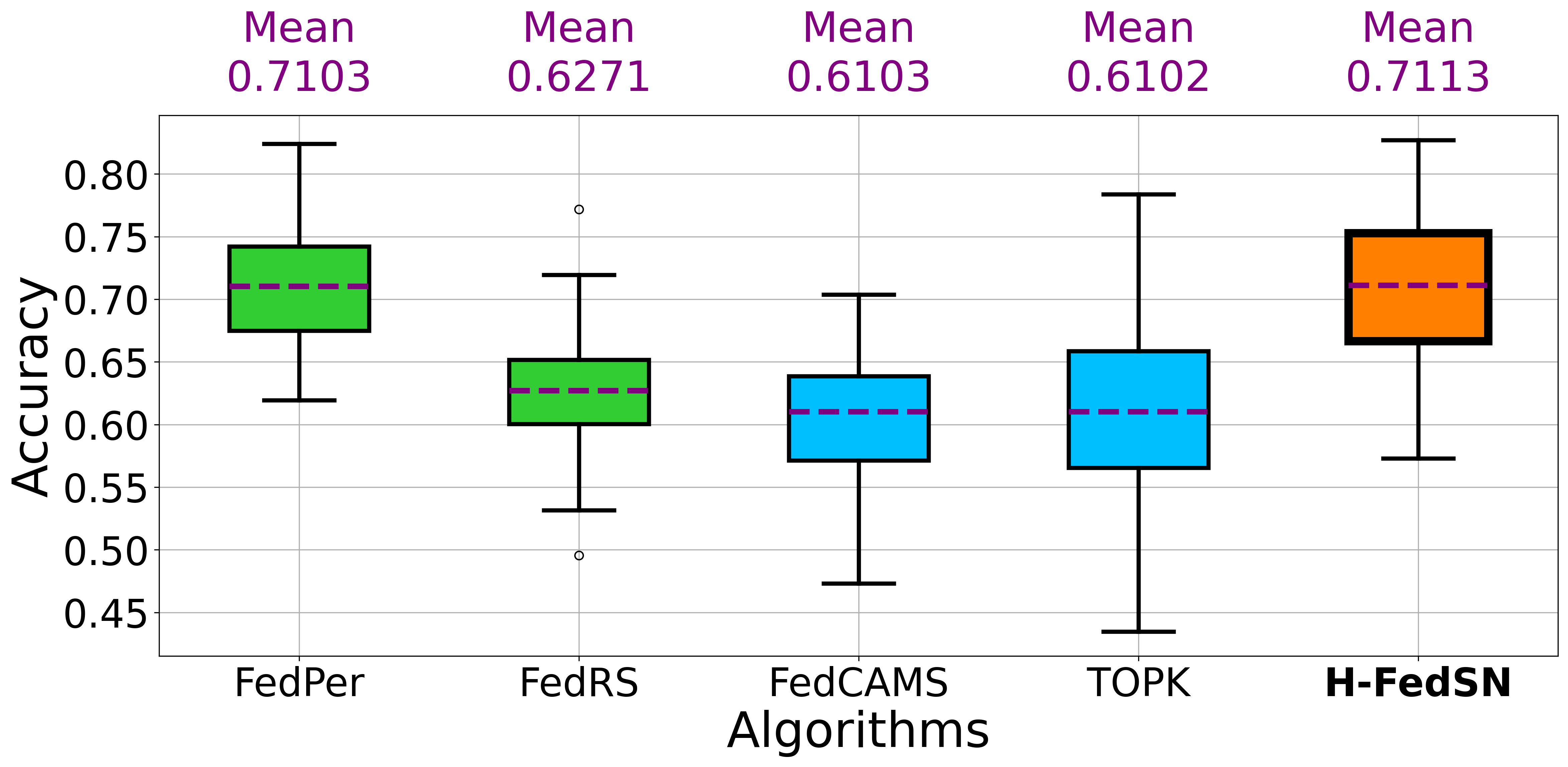} }}
    \subfloat[WIDAR]{{\includegraphics[width=0.32\linewidth]{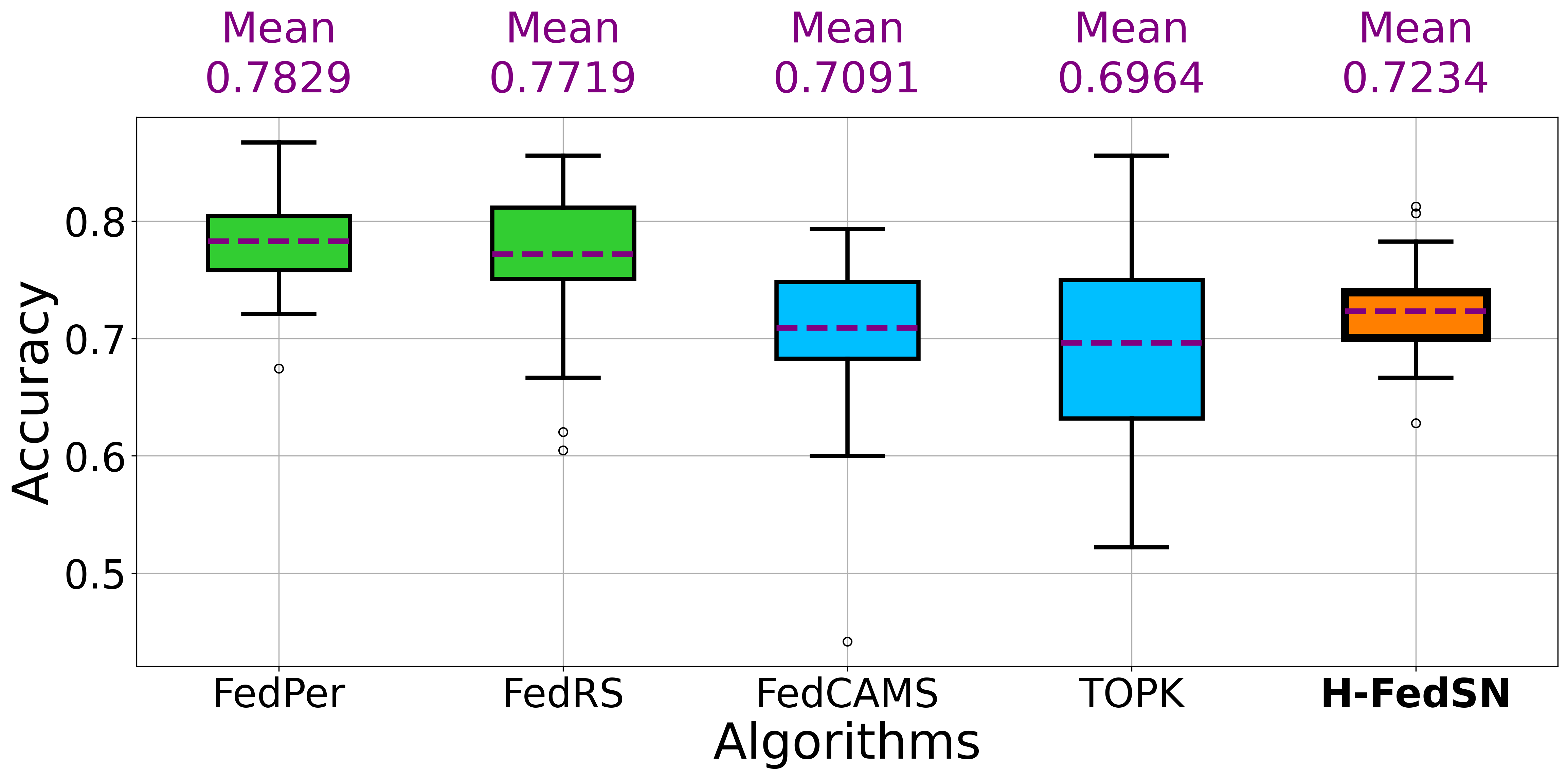} }}
    \subfloat[WISDM(PHONE)]{{\includegraphics[width=0.33\linewidth]{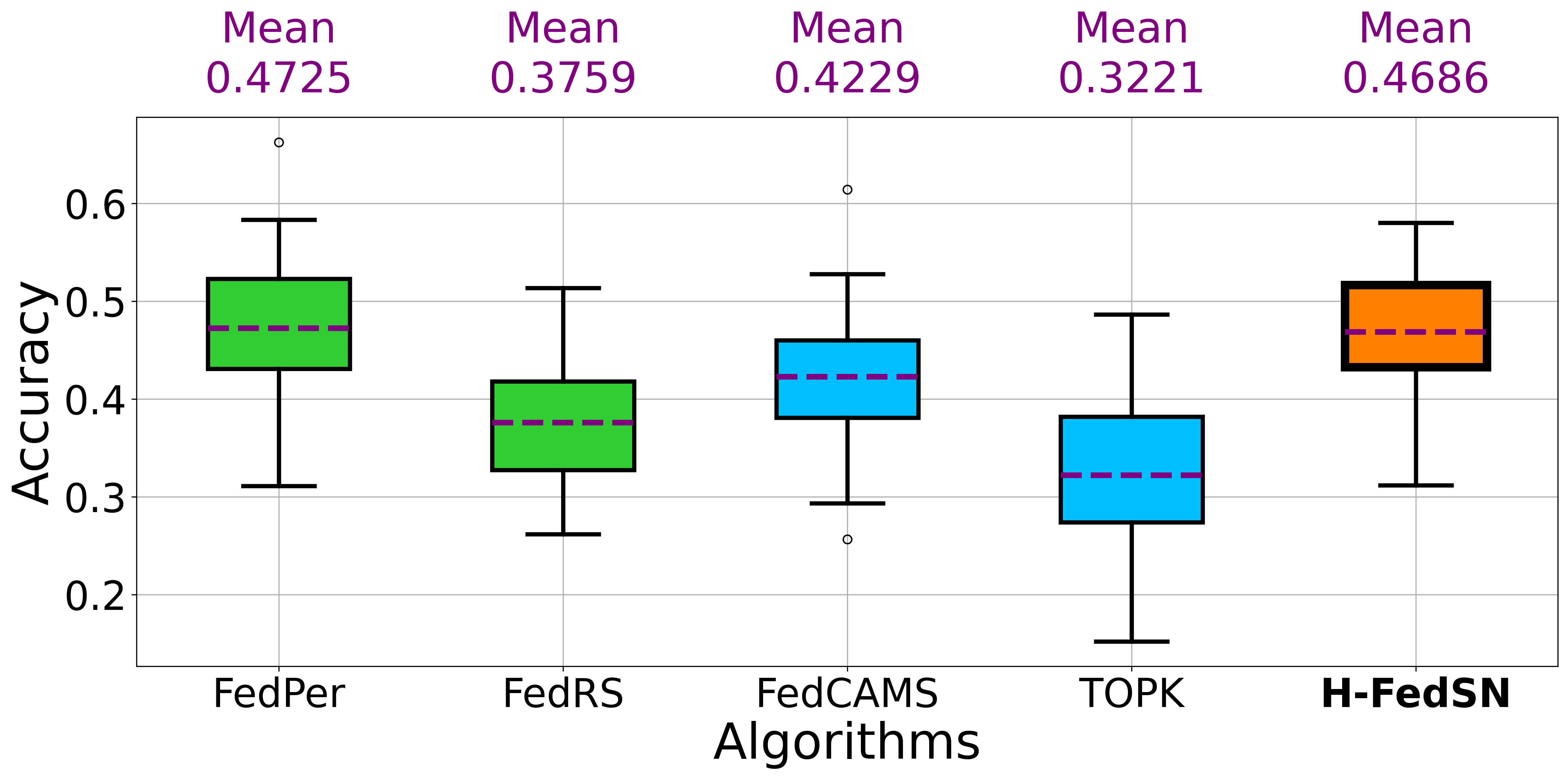} }}
    \caption{Inference accuracy for different algorithms across different datasets in \textbf{Imbalanced E5C50} configuration.}
    \label{fig:im_e5_c50_inf_acc} 
\end{figure}
% As shown in Fig.~\ref{fig:comm} and Fig.~\ref{fig:im_e5_c50_inf_acc}, \methodname{} achieves accuracy levels comparable to baseline methods while simultaneously minimizing communication overhead. This consistent performance is attributed to the strategic use of personalized layers and Bayesian aggregation, which effectively address heterogeneity and imbalance. By employing personalized layers and efficient masking techniques, the amount of model parameters exchanged between clients is significantly reduced, leading to the lowest communication costs among all evaluated methods. This ability to minimize communication overhead while preserving high accuracy highlights \methodname{}'s advantages in addressing the challenges of device/data imbalance and communication efficiency in real-world distributed learning environments.
%highlight
As shown in Fig.~\ref{fig:comm} and Fig.~\ref{fig:im_e5_c50_inf_acc}, on the wisdm datasets, \methodname{} matches the strongest personalization baseline FedPer (WISDM(WATCH): 71.13\% versus 71.03\%; WISDM(PHONE): 46.86\% versus 47.25\%). On WIDAR, FedPer (78.29\%) and FedRS (77.19\%) lead, with \methodname{} at 72.34\%. Across all three datasets, \methodname{} transmits roughly two orders of magnitude less per round than the personalization baselines and several times less than FedCAMS and TOPK, and surpasses both communication-optimized baselines on accuracy and per-client consistency. The personalized layers retained locally and the Bayesian aggregation propagating only sparse binary masks across the hierarchy together absorb the uneven client distribution.
%highlight

\subsubsection{E20C50 Configuration}

In the \textbf{E20C50} configuration, \methodname{} maintains minimal communication costs while achieving strong inference accuracy across clients. 
\begin{figure}[h!]
    \centering
    % \subfloat[MNIST]{{\includegraphics[width=0.5\linewidth]{exp_res/e20_c50/e20-c50-MNIST.png} }}
    \subfloat[WISDM(WATCH)]{{\includegraphics[width=0.33\linewidth]{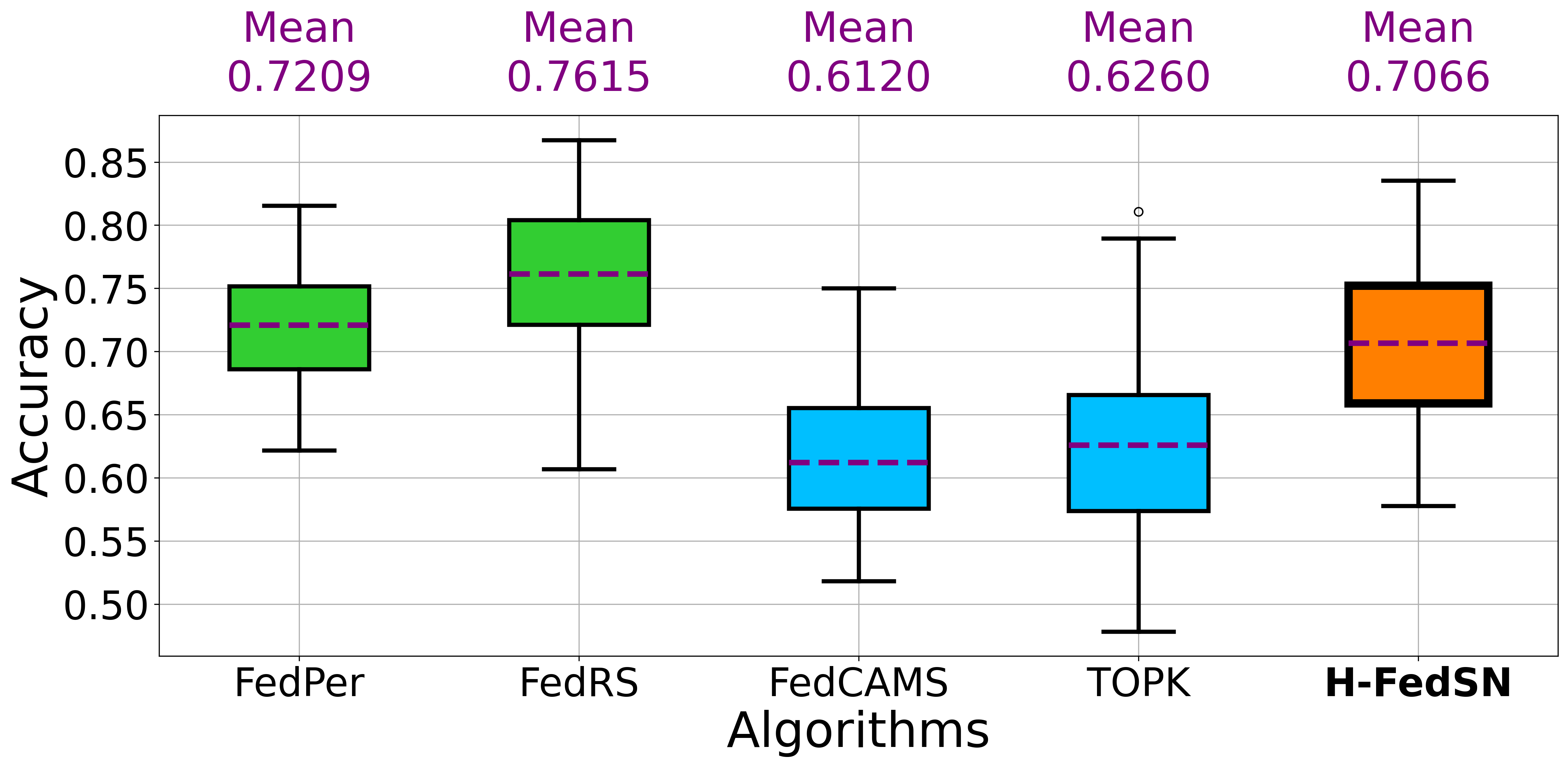} }}
    \subfloat[WIDAR]{{\includegraphics[width=0.32\linewidth]{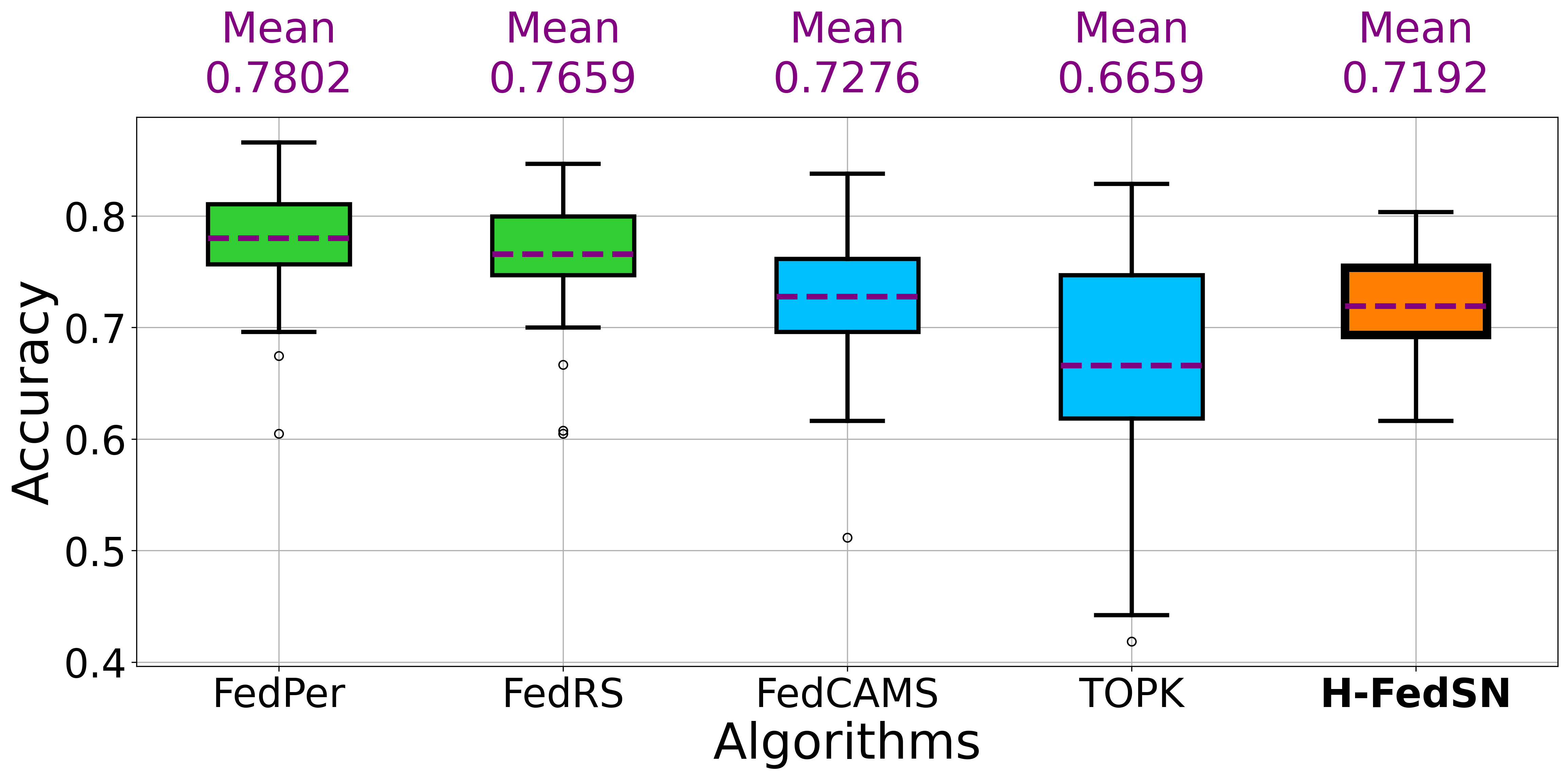} }}
    \subfloat[WISDM(PHONE)]{{\includegraphics[width=0.33\linewidth]{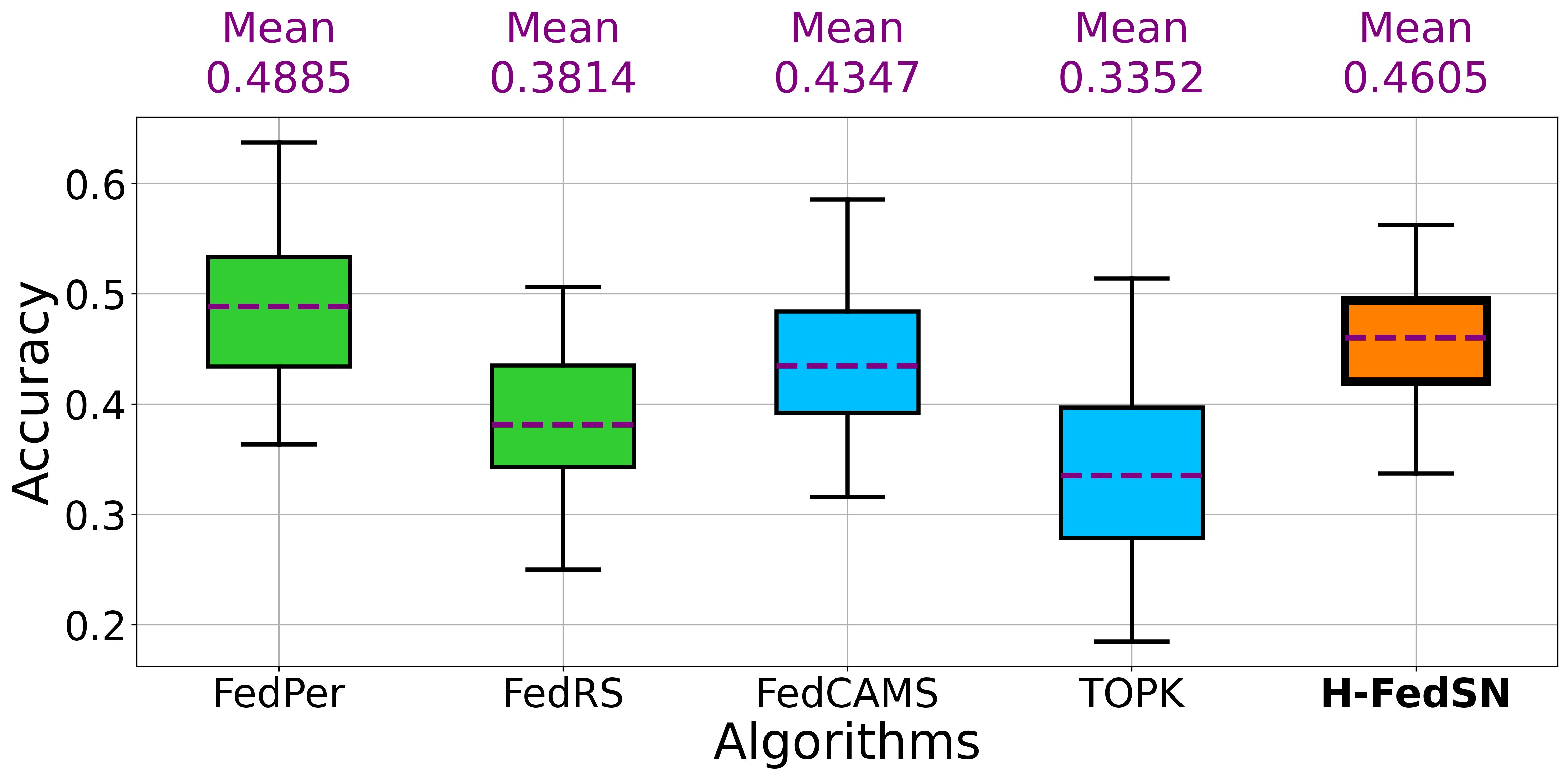} }}
    \caption{Inference accuracy for different algorithms across different datasets in \textbf{E20C50} configuration.}
    \label{fig:e20_c50_inf_acc} 
\end{figure}

% As shown in Fig.\ref{fig:comm}, \methodname{} significantly reduces communication expenses compared to the baselines, despite operating at a larger network scale. This reduction stems from the sparse-mask mechanism, which effectively compresses model updates without compromising performance. Furthermore, in Fig.\ref{fig:e20_c50_inf_acc}, we illustrate not just the mean accuracy but also the distribution of accuracies among all clients. Notably, \methodname{}’s accuracies are not only high on average but also concentrated within a narrower range. This reduced variance indicates that \methodname{} offers more consistent performance across heterogeneous clients, minimizing the risk of poorly performing outlier devices. By contrast, the baseline methods exhibit wider dispersion in their client-level accuracies, suggesting that some clients may converge to suboptimal models. Overall, \methodname{} demonstrates both robustness and scalability under the E20C50 scenario, providing higher and more uniform accuracy at substantially lower communication cost.
%highlight
As shown in Fig.\ref{fig:e20_c50_inf_acc} and Fig.\ref{fig:comm}, \methodname{} stays close to these personalization baselines (FedPer, FedRS) but transmits roughly two orders of magnitude less per round. The communication-optimized baselines (FedCAMS, TOPK) trail \methodname{} on every dataset both in mean accuracy and per-client consistency. \methodname{} occupies the low-communication corner of the accuracy--communication Pareto frontier, achieving the lowest transmission cost while staying within a small accuracy margin of the highest-accuracy methods.
%highlight

\subsubsection{E2C50 Configuration}
\begin{figure}[h!]
    \centering
    % \subfloat[MNIST]{{\includegraphics[width=0.25\linewidth]{exp_res/e2_c50/e2-c50-MNIST.png} }}
    \subfloat[WISDM(WATCH)]{{\includegraphics[width=0.33\linewidth]{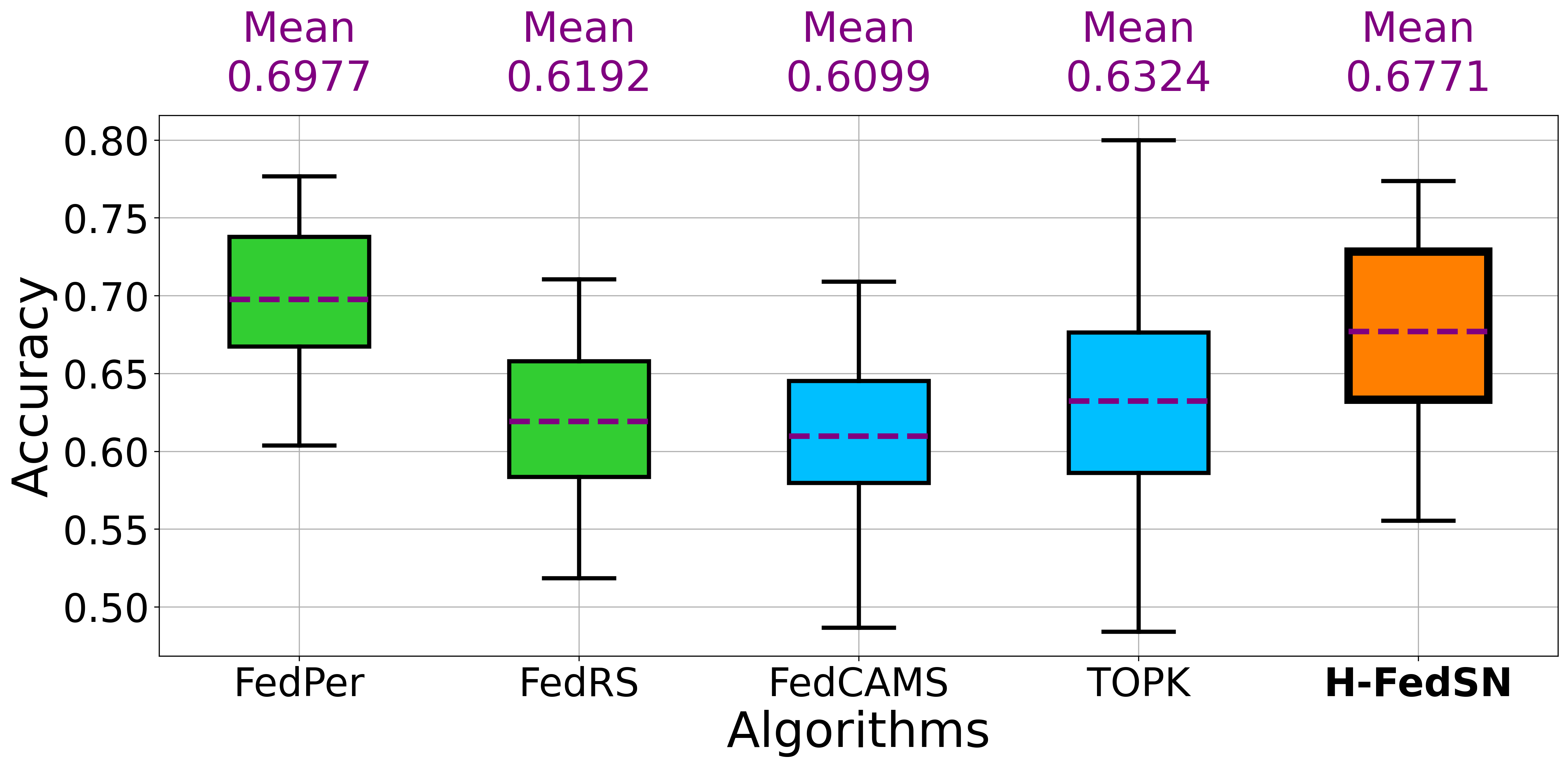} }}
    \subfloat[WIDAR]{{\includegraphics[width=0.32\linewidth]{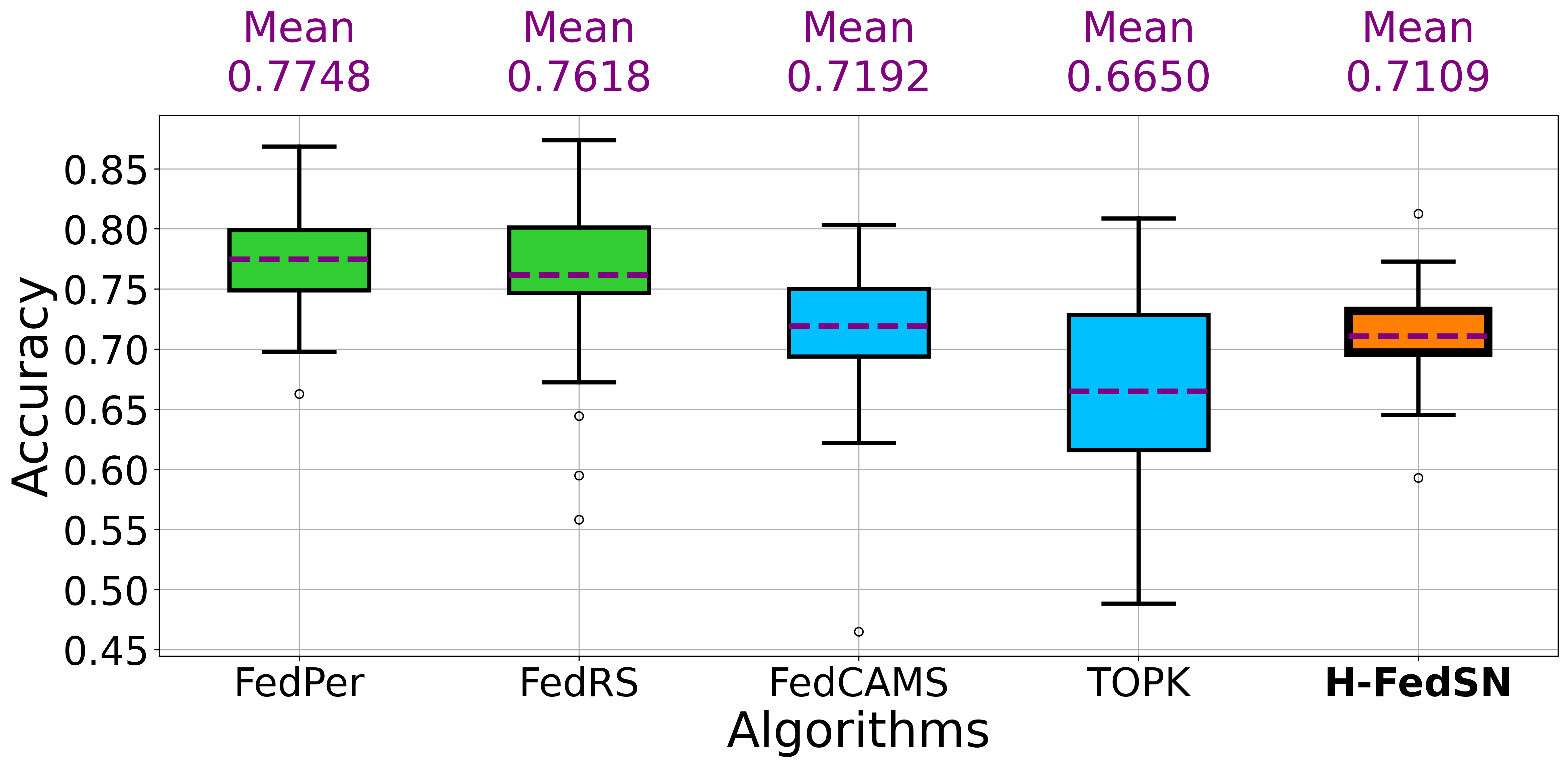} }}
    \subfloat[WISDM(PHONE)]{{\includegraphics[width=0.33\linewidth]{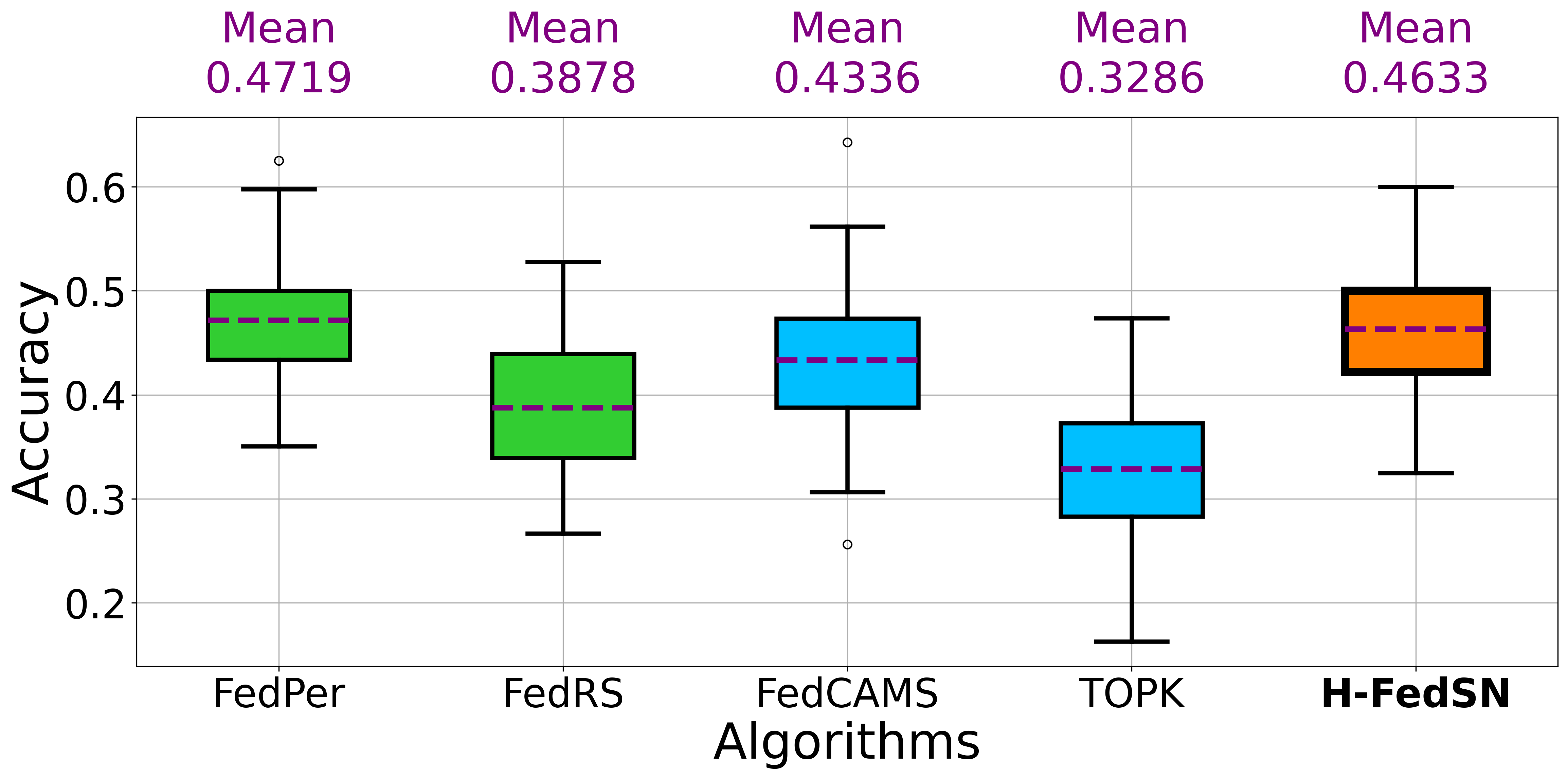} }}
    \caption{Inference accuracy for different algorithms across different datasets in \textbf{E2C50} configuration.}
    \label{fig:e2_c50_inf_acc} 
\end{figure}
% The \textbf{E2C50} configuration, where each edge server coordinates a high density of clients, places a considerable strain on network resources. As shown in Fig.\ref{fig:comm}, \methodname{} transmits the smallest overall volume of model parameters compared with the other baselines, resulting in the lowest communication cost despite the intense client load. Furthermore, in Fig.\ref{fig:e2_c50_inf_acc}, we observe that \methodname{} not only maintains a competitive mean accuracy but also achieves a noticeably smaller variance across clients. This smaller spread in performance indicates that \methodname{} better accommodates heterogeneous data at scale, preventing certain clients from converging to substantially weaker models. Consequently, \methodname{} effectively addresses both the bandwidth limitations and accuracy distribution challenges in high-density federated environments.
%highlight
The \textbf{E2C50} configuration, where each edge server coordinates a high density of clients, places a considerable strain on network resources. In Fig.\ref{fig:e2_c50_inf_acc}, FedPer reaches the highest mean accuracy on every dataset, and \methodname{} stays close to these personalization baselines. FedCAMS and TOPK fall below \methodname{} on every dataset. Per-round transmission cost of \methodname{} (Fig.~\ref{fig:comm}) stays roughly two orders of magnitude below FedPer and FedRS and several times below FedCAMS and TOPK.
%highlight

\begin{table}[htbp]
    \centering
    \caption{Accuracy on MNIST Across Experiment Configurations (E5C50, imbalanced E2C50, E20C50)}
    \scalebox{0.9}{
    \setlength{\tabcolsep}{3pt} 
    \renewcommand{\arraystretch}{1.0} 
    \begin{tabular}{@{}lllr@{}} 
        \toprule
        \textbf{Setting} & \textbf{FedCAMS} & \textbf{TOPK} & \textbf{H-FedSN} \\
        \midrule
        Imbalanced E5C50 & 0.9956 & 0.9936 & 0.9893  \\
        E2C50 &  0.9964 & 0.9941 & 0.9878  \\
        E20C50 &  0.9967 & 0.9960 & 0.9908 \\
        \bottomrule
    \end{tabular}}
    \label{tab:mnist_extended}
\end{table}

Although we also evaluate the methods on \textbf{MNIST} in the subsequent E2C50, imbalanced E5C50, and E20C50 configurations, their accuracies differ by less than 1\% across all baselines, offering little additional insight beyond what is already established under E2C5. As summarized in Table~\ref{tab:mnist_extended}, all methods reach near-identical accuracy on MNIST, while \methodname{} maintains the lowest communication overhead. Hence, we omit dedicated MNIST accuracy curves in the following figures to avoid redundancy, but we include this table to confirm that the same efficiency advantage of \methodname{} holds in those larger-scale or imbalanced settings.

% We also evaluated \methodname{} on two additional configurations: E20C50 
% and E2C50. These results demonstrate similar patterns of communication 
% efficiency and accuracy maintenance. Detailed analyses of these configurations are provided in 
% Appx. B.

% In summary, \methodname{} achieves considerable reductions in communication cost and maintains high inference accuracy compared to all baselines. 
% % Compared to specific personalization approaches (e.g., FedPer, FedRS), \methodname{} offers similar accuracy at much lower communication overhead. Moreover, compared to communication-focused methods (e.g., FedCAMS, TOPK), \methodname{} both transmit less data and yields higher accuracy.
% These findings confirm the method’s efficiency and efficacy in hierarchical federated learning scenarios.
In summary, \methodname{} effectively balances communication efficiency and inference accuracy, consistently outperforming baseline methods across four IoT scenarios. Its personalized sparse structure and Bayesian aggregation strategy significantly reduce communication overhead while maintaining robust performance under data heterogeneity and imbalanced client distributions. These results highlight the practical advantages of \methodname{} for IoT-oriented hierarchical federated learning.

\section{Conclusion}

In this paper, we propose \methodname{}, a novel hierarchical federated learning method designed specifically to address the dual challenges of communication efficiency and inference accuracy in practical IoT environments. By leveraging compact binary masks to achieve network sparsification, retaining personalized model parameters locally at each client, and employing Bayesian aggregation at both edge and cloud layers, \methodname{} effectively mitigates issues arising from data heterogeneity, imbalanced client participation, and bandwidth constraints. Experimental evaluations demonstrate that \methodname{} achieves comparable or superior inference accuracy relative to state-of-the-art baselines, while simultaneously reducing communication costs by one to two orders of magnitude.
% For future work, we plan to further explore the scalability of \methodname{} by extending it to larger-scale hierarchical scenarios involving more than three aggregation layers, as well as investigating adaptive mechanisms to dynamically adjust sparsity levels according to real-time network conditions and client capabilities.
For future work, we plan to extend \methodname{} along several directions: scaling to larger hierarchical scenarios with more than three aggregation layers, designing adaptive mechanisms that dynamically adjust sparsity levels under varying network conditions, exploring alternative mask parameterizations and broader Bayesian prior designs.

\section{Acknowledgements}
This research is sponsored by funding from Stanford’s Center for Sustainable Development and Global Competitiveness (SDGC) and the Yonghua Foundation.

\bibliographystyle{ACM-Reference-Format}
\bibliography{sample-base}

@article{yang2019federated,
  title={Federated machine learning: Concept and applications},
  author={Yang, Qiang and Liu, Yang and Chen, Tianjian and Tong, Yongxin},
  journal={ACM Transactions on Intelligent Systems and Technology (TIST)},
  volume={10},
  number={2},
  pages={1--19},
  year={2019},
  publisher={ACM New York, NY, USA}
}

@article{li2020federated,
  title={Federated optimization in heterogeneous networks},
  author={Li, Tian and Sahu, Anit Kumar and Zaheer, Manzil and Sanjabi, Maziar and Talwalkar, Ameet and Smith, Virginia},
  journal={Proceedings of Machine learning and systems},
  volume={2},
  pages={429--450},
  year={2020}
}

@article{anthopoulos2015understanding,
  title={Understanding the smart city domain: A literature review},
  author={Anthopoulos, Leonidas G},
  journal={Transforming city governments for successful smart cities},
  pages={9--21},
  year={2015},
  publisher={Springer}
}

@inproceedings{mcmahan2017communication,
  title={Communication-efficient learning of deep networks from decentralized data},
  author={McMahan, Brendan and Moore, Eider and Ramage, Daniel and Hampson, Seth and y Arcas, Blaise Aguera},
  booktitle={Artificial intelligence and statistics},
  pages={1273--1282},
  year={2017},
  organization={PMLR}
}

@inproceedings{karimireddy2020scaffold,
  title={Scaffold: Stochastic controlled averaging for federated learning},
  author={Karimireddy, Sai Praneeth and Kale, Satyen and Mohri, Mehryar and Reddi, Sashank and Stich, Sebastian and Suresh, Ananda Theertha},
  booktitle={International conference on machine learning},
  pages={5132--5143},
  year={2020},
  organization={PMLR}
}

@inproceedings{liu2020client,
  title={Client-edge-cloud hierarchical federated learning},
  author={Liu, Lumin and Zhang, Jun and Song, SH and Letaief, Khaled B},
  booktitle={ICC 2020-2020 IEEE international conference on communications (ICC)},
  pages={1--6},
  year={2020},
  organization={IEEE}
}

@article{Sheng2023Federated,title={Federated Learning Technology in Serial Topology for IoT Networks},author={Jianhang Sheng and Jian Xiong and Bo Liu},journal={2023 IEEE International Symposium on Broadband Multimedia Systems and Broadcasting (BMSB)},year={2023},pages={1-5},doi={10.1109/BMSB58369.2023.10211485}}

@article{aji2017sparse,
  title={Sparse communication for distributed gradient descent},
  author={Aji, Alham Fikri and Heafield, Kenneth},
  journal={arXiv preprint arXiv:1704.05021},
  year={2017}
}

@article{Imteaj2021A,title={A Survey on Federated Learning for Resource-Constrained IoT Devices},author={Ahmed Imteaj and Urmish Thakker and Shiqiang Wang and Jian Li and M. Amini},journal={IEEE Internet of Things Journal},year={2021},volume={9},pages={1-24},doi={10.1109/jiot.2021.3095077}}

@article{Ni2023Semi-Federated,title={Semi-Federated Learning for Collaborative Intelligence in Massive IoT Networks},author={Wanli Ni and Jingheng Zheng and Hui Tian},journal={IEEE Internet of Things Journal},year={2023},volume={10},pages={11942-11943},doi={10.1109/JIOT.2023.3253853}}

@article{Khan2020Federated,title={Federated Learning for Internet of Things: Recent Advances, Taxonomy, and Open Challenges},author={L. U. Khan and W. Saad and Zhu Han and E. Hossain and C. Hong},journal={IEEE Communications Surveys \& Tutorials},year={2020},volume={23},pages={1759-1799},doi={10.1109/COMST.2021.3090430}}

@article{Wu2020Personalized,title={Personalized Federated Learning for Intelligent IoT Applications: A Cloud-Edge Based Framework},author={Qiong Wu and Kaiwen He and Xu Chen},journal={IEEE Open Journal of the Computer Society},year={2020},doi={10.1109/OJCS.2020.2993259}}

@article{bengio2013estimating,
  title={Estimating or propagating gradients through stochastic neurons for conditional computation},
  author={Bengio, Yoshua and L{\'e}onard, Nicholas and Courville, Aaron},
  journal={arXiv preprint arXiv:1308.3432},
  year={2013}
}

@article{zhou2019deconstructing,
  title={Deconstructing lottery tickets: Zeros, signs, and the supermask},
  author={Zhou, Hattie and Lan, Janice and Liu, Rosanne and Yosinski, Jason},
  journal={Advances in neural information processing systems},
  volume={32},
  year={2019}
}

@article{ferreira2021bayesian,
  title={Bayesian SignSGD Optimizer for Federated Learning},
  author={Ferreira, Paulo Abelha and da Silva, Pablo Nascimento and Gottin, Vinicius and Stelling, Roberto and Calmon, Tiago},
  journal={Advances in Neural Information Processing Systems},
  volume={34},
  year={2021}
}

@article{Wang2022Communication-Efficient,title={Communication-Efficient Adaptive Federated Learning},author={Yujia Wang and Lu Lin and Jinghui Chen},journal={ArXiv},year={2022},volume={abs/2205.02719},doi={10.48550/arXiv.2205.02719}}

@article{arivazhagan2019federated,
  title={Federated learning with personalization layers},
  author={Arivazhagan, Manoj Ghuhan and Aggarwal, Vinay and Singh, Aaditya Kumar and Choudhary, Sunav},
  journal={arXiv preprint arXiv:1912.00818},
  year={2019}
}

@inproceedings{10.1145/3447548.3467254,
author = {Li, Xin-Chun and Zhan, De-Chuan},
title = {FedRS: Federated Learning with Restricted Softmax for Label Distribution Non-IID Data},
year = {2021},
isbn = {9781450383325},
publisher = {Association for Computing Machinery},
address = {New York, NY, USA},
url = {https://doi.org/10.1145/3447548.3467254},
doi = {10.1145/3447548.3467254},
booktitle = {Proceedings of the 27th ACM SIGKDD Conference on Knowledge Discovery \& Data Mining},
pages = {995–1005},
numpages = {11},
location = {Virtual Event, Singapore},
series = {KDD '21}
}

@article{fallah2020personalized,
  title={Personalized federated learning: A meta-learning approach},
  author={Fallah, Alireza and Mokhtari, Aryan and Ozdaglar, Asuman},
  journal={arXiv preprint arXiv:2002.07948},
  year={2020}
}

@article{lecun1998gradient,
  title={Gradient-based learning applied to document recognition},
  author={LeCun, Yann and Bottou, L{\'e}on and Bengio, Yoshua and Haffner, Patrick},
  journal={Proceedings of the IEEE},
  volume={86},
  number={11},
  pages={2278--2324},
  year={1998},
  publisher={Ieee}
}

@inproceedings{lockhart2011design,
  title={Design considerations for the WISDM smart phone-based sensor mining architecture},
  author={Lockhart, Jeffrey W and Weiss, Gary M and Xue, Jack C and Gallagher, Shaun T and Grosner, Andrew B and Pulickal, Tony T},
  booktitle={Proceedings of the Fifth International Workshop on Knowledge Discovery from Sensor Data},
  pages={25--33},
  year={2011}
}

@article{weiss2019smartphone,
  title={Smartphone and smartwatch-based biometrics using activities of daily living},
  author={Weiss, Gary M and Yoneda, Kenichi and Hayajneh, Thaier},
  journal={Ieee Access},
  volume={7},
  pages={133190--133202},
  year={2019},
  publisher={IEEE}
}

@data{widardata2020,
doi = {10.21227/7znf-qp86},
url = {https://dx.doi.org/10.21227/7znf-qp86},
author = {Yang, Zheng and Zhang, Yi and Zhang, Guidong and Zheng, Yue and Chi, Guoxuan},
publisher = {IEEE Dataport},
title = {Widar 3.0: WiFi-based Activity Recognition Dataset},
year = {2020} }

@inproceedings{zheng2019zero,
  title={Zero-effort cross-domain gesture recognition with Wi-Fi},
  author={Zheng, Yue and Zhang, Yi and Qian, Kun and Zhang, Guidong and Liu, Yunhao and Wu, Chenshu and Yang, Zheng},
  booktitle={Proceedings of the 17th annual international conference on mobile systems, applications, and services},
  pages={313--325},
  year={2019}
}

@ARTICLE{9303442,
  author={Wu, Xueyu and Yao, Xin and Wang, Cho-Li},
  journal={IEEE Transactions on Parallel and Distributed Systems}, 
  title={FedSCR: Structure-Based Communication Reduction for Federated Learning}, 
  year={2021},
  volume={32},
  number={7},
  pages={1565-1577},
  doi={10.1109/TPDS.2020.3046250}}

@article{Nori2021Fast,title={Fast Federated Learning by Balancing Communication Trade-Offs},author={M. Nori and Sangseok Yun and Il Kim},journal={IEEE Transactions on Communications},year={2021},volume={69},pages={5168-5182},doi={10.1109/TCOMM.2021.3083316}}

@article{Ren2022Toward,title={Toward Communication-Learning Trade-Off for Federated Learning at the Network Edge},author={Jian-ji Ren and Wanli Ni and Hui Tian},journal={IEEE Communications Letters},year={2022},volume={26},pages={1858-1862},doi={10.1109/LCOMM.2022.3174295}}

@inproceedings{ravi2005activity,
  title={Activity recognition from accelerometer data},
  author={Ravi, Nishkam and Dandekar, Nikhil and Mysore, Preetham and Littman, Michael L},
  booktitle={Aaai},
  volume={5},
  number={2005},
  pages={1541--1546},
  year={2005},
  organization={Pittsburgh, PA}
}

@article{reyes2016transition,
  title={Transition-aware human activity recognition using smartphones},
  author={Reyes-Ortiz, Jorge-L and Oneto, Luca and Sam{\`a}, Albert and Parra, Xavier and Anguita, Davide},
  journal={Neurocomputing},
  volume={171},
  pages={754--767},
  year={2016},
  publisher={Elsevier}
}

@article{ronao2016human,
  title={Human activity recognition with smartphone sensors using deep learning neural networks},
  author={Ronao, Charissa Ann and Cho, Sung-Bae},
  journal={Expert systems with applications},
  volume={59},
  pages={235--244},
  year={2016},
  publisher={Elsevier}
}

@inproceedings{li2022federated,
  title={Federated learning on non-iid data silos: An experimental study},
  author={Li, Qinbin and Diao, Yiqun and Chen, Quan and He, Bingsheng},
  booktitle={2022 IEEE 38th international conference on data engineering (ICDE)},
  pages={965--978},
  year={2022},
  organization={IEEE}
}

@inproceedings{
liu2024fedbcgd,
title={Fed{BCGD}: Communication-Efficient Accelerated Block Coordinate Gradient Descent for Federated Learning},
author={Junkang Liu and Fanhua Shang and Yuanyuan Liu and Hongying Liu and Yuangang Li and YunXiang Gong},
booktitle={ACM Multimedia 2024},
year={2024},
url={https://openreview.net/forum?id=lAFO0SUjXD}
}

@INPROCEEDINGS{10335537EAFL,
  author={Sun, Yu and Gao, Xin and Shen, Ying and Xie, Jianfei and Yang, Jiandong and Si, Nong},
  booktitle={2023 IEEE 3rd International Conference on Computer Systems (ICCS)}, 
  title={A model parameter update strategy for enhanced asynchronous federated learning algorithm}, 
  year={2023},
  volume={},
  number={},
  pages={9-15},
  doi={10.1109/ICCS59700.2023.10335537}}

@ARTICLE{9288933FTTQ,
  author={Xu, Jinjin and Du, Wenli and Jin, Yaochu and He, Wangli and Cheng, Ran},
  journal={IEEE Transactions on Neural Networks and Learning Systems}, 
  title={Ternary Compression for Communication-Efficient Federated Learning}, 
  year={2022},
  volume={33},
  number={3},
  pages={1162-1176},
  doi={10.1109/TNNLS.2020.3041185}}

@article{Virk2020Smart,
title={Smart Farming: An Overview},
author={A. L. Virk and Mehmood Ali Noor and S. Fiaz and S. Hussain and H. Hussain and M. Rehman and M. Ahsan and Wei Ma},
journal={Smart Village Technology},
year={2020},
doi={10.1007/978-3-030-37794-6_10}
}

@INPROCEEDINGS{9533879,
  author={Tijani, Saheed A. and Ma, Xingjun and Zhang, Ran and Jiang, Frank and Doss, Robin},
  booktitle={2021 International Joint Conference on Neural Networks (IJCNN)}, 
  title={Federated Learning with Extreme Label Skew: A Data Extension Approach}, 
  year={2021},
  volume={},
  number={},
  pages={1-8},
  doi={10.1109/IJCNN52387.2021.9533879}}

@INPROCEEDINGS{8066704,
  author={Nguyen, Hoa Hong and Mirza, Farhaan and Naeem, M. Asif and Nguyen, Minh},
  booktitle={2017 IEEE 21st International Conference on Computer Supported Cooperative Work in Design (CSCWD)}, 
  title={A review on IoT healthcare monitoring applications and a vision for transforming sensor data into real-time clinical feedback}, 
  year={2017},
  volume={},
  number={},
  pages={257-262},
  doi={10.1109/CSCWD.2017.8066704}}

@article{purohit2021evaluation,
  title={Evaluation of the efficacy of IoT deployment on Petro-Retail Operations},
  author={Purohit, Santanu and others},
  journal={Turkish Journal of Computer and Mathematics Education (TURCOMAT)},
  volume={12},
  number={4},
  pages={356--363},
  year={2021}
}

@article{Ma_2024, title={Personalized client-edge-cloud hierarchical federated learning in mobile edge computing}, volume={13}, ISSN={2192-113X}, url={http://dx.doi.org/10.1186/s13677-024-00721-w}, DOI={10.1186/s13677-024-00721-w}, number={1}, journal={Journal of Cloud Computing}, publisher={Springer Science and Business Media LLC}, author={Ma, Chunmei and Li, Xiangqian and Huang, Baogui and Li, Guangshun and Li, Fengyin}, year={2024}, month=dec }

@ARTICLE{khan2021federated,
  author={Khan, Latif U. and Saad, Walid and Han, Zhu and Hossain, Ekram and Hong, Choong Seon},
  journal={IEEE Communications Surveys \& Tutorials}, 
  title={Federated Learning for Internet of Things: Recent Advances, Taxonomy, and Open Challenges}, 
  year={2021},
  volume={23},
  number={3},
  pages={1759-1799},
  doi={10.1109/COMST.2021.3090430}}

@article{zhang2022federated,
  title={Federated learning for the internet of things: Applications, challenges, and opportunities},
  author={Zhang, Tuo and Gao, Lei and He, Chaoyang and Zhang, Mi and Krishnamachari, Bhaskar and Avestimehr, A Salman},
  journal={IEEE Internet of Things Magazine},
  volume={5},
  number={1},
  pages={24--29},
  year={2022},
  publisher={IEEE}
}

@ARTICLE{shi2016edge,
  author={Shi, Weisong and Cao, Jie and Zhang, Quan and Li, Youhuizi and Xu, Lanyu},
  journal={IEEE Internet of Things Journal}, 
  title={Edge Computing: Vision and Challenges}, 
  year={2016},
  volume={3},
  number={5},
  pages={637-646},
  doi={10.1109/JIOT.2016.2579198}}

@ARTICLE{satyanarayanan2017emergence,
  author={Satyanarayanan, Mahadev},
  journal={Computer}, 
  title={The Emergence of Edge Computing}, 
  year={2017},
  volume={50},
  number={1},
  pages={30-39},
  doi={10.1109/MC.2017.9}}

@article{quy2022smart,
  title={Smart healthcare IoT applications based on fog computing: architecture, applications and challenges},
  author={Quy, Vu Khanh and Hau, Nguyen Van and Anh, Dang Van and Ngoc, Le Anh},
  journal={Complex \& Intelligent Systems},
  volume={8},
  number={5},
  pages={3805--3815},
  year={2022},
  publisher={Springer}
}

\clearpage

\appendix
\section*{Supplementary Material}

\setcounter{section}{0}
\setcounter{figure}{0}
\setcounter{table}{0}
\makeatletter 
\renewcommand{\thesection}{\Alph{section}}
\renewcommand{\theHsection}{\Alph{section}}
\renewcommand{\thefigure}{A\arabic{figure}} % Figure counter representation
\renewcommand{\theHfigure}{A\arabic{figure}} % Hyperref figure hyperlink hook
\renewcommand{\thetable}{A\arabic{table}}
\renewcommand{\theHtable}{A\arabic{table}}
\makeatother

\renewcommand{\thetable}{A\arabic{table}}
\setcounter{equation}{0}
\renewcommand{\theequation}{A\arabic{equation}}

\section{Experimental Details}
\subsection{Sensitivity to Source-Paper Learning Rates}
\label{appx:lr}

To examine whether the unified baseline learning rate(LR) used in the main experiments under-optimizes the comparison methods, we conducted an additional sensitivity study using the learning rates reported in the original baseline papers where directly applicable. All other experimental settings were kept fixed, same setting with \S~\ref{exp:exp_setting}, and using the E2C5 setting.

Table~\ref{tab:source_lr_sensitivity} reports the accuracy of each baseline when using its source-paper learning rate. Values in parentheses denote the absolute percentage-point change relative to the corresponding result with the unified learning rate \(10^{-3}\) used in the main experiments.

\begin{table}[htbp]
\centering
\caption{Baseline sensitivity to source-paper learning rates under E2C5 setting. Parentheses show percentage-point changes from \(10^{-3}\).}
\label{tab:source_lr_sensitivity}
\scriptsize
\setlength{\tabcolsep}{3pt}
\begin{tabular}{lcccc}
\toprule
\textbf{Method} & \textbf{source-paper LR} & \textbf{WISDM(WATCH) } & \textbf{WIDAR} & \textbf{WISDM(PHONE)} \\
\midrule
HierFAVG & 0.01  & 45.16 (-28.66) & 44.66 (-33.23) & 26.75 (-10.32) \\
FedPer   & 0.01  & 57.85 (-14.61) & 44.66 (-38.19) & 44.13 (-4.23)  \\
FedRS    & 0.03  & 13.56 (-62.41) & 30.80 (-43.52) & 16.71 (-21.43) \\
FedCAMS  & 0.01  & 18.14 (-41.51) & 39.18 (-31.79) & 21.17 (-18.50) \\
TOPK     & 0.005 & 44.83 (-14.94) & 50.96 (-11.22) & 27.64 (-2.38)  \\
\bottomrule
\end{tabular}
\end{table}

The source-paper learning rates do not improve baseline performance in this setting. Instead, they reduce accuracy across all tested baselines and datasets, with especially large drops for FedRS and FedCAMS. This suggests that learning rates tuned in the original baseline settings do not directly transfer to the three-tier HFL setting with IoT sensor data and repeated edge-level aggregation. The unified learning rate \(10^{-3}\) used in the main experiments therefore provides a more stable baseline configuration under our experimental protocol.

\subsection{Ablation on Historical Window Size}
\label{sec:supp_window_ablation}

We analyze the sensitivity of \methodname{} to the historical window size $w$, the reset interval of the Bayesian $\alpha/\beta$ priors at the edge and cloud levels. To do so, we vary $w$ over $\{1, 5, 10, 20, 50\}$ and evaluate the resulting accuracy on WIDAR under the E2C5 setting, the highest-dimensional IoT dataset in our benchmark (22 channels of $20\times 20$ WiFi CSI patterns).

\begin{table}[htbp]
\centering
\caption{Sensitivity of \methodname{} to the historical window size $w$ on WIDAR. The default $w=10$ used in the main paper is marked with $\dagger$. E2C5 setting.}
\label{tab:supp_window_ablation}
\scriptsize
\setlength{\tabcolsep}{3pt}
\begin{tabular}{l c c c c c}
\toprule
\textbf{$w$} & \textbf{1} & \textbf{5} & \textbf{10}$^{\dagger}$ & \textbf{20} & \textbf{50} \\
\midrule
WIDAR accuracy & 66.06\% & 75.23\% & \textbf{76.37\%} & 74.11\% & 72.98\% \\
\bottomrule
\end{tabular}
\end{table}

Table~\ref{tab:supp_window_ablation} summarizes the results. Overall, \methodname{} achieves the strongest accuracy at $w=10$, which we therefore use in all experiments. Smaller windows reset the priors before a useful posterior accumulates and reduce the Bayesian aggregation to single-round voting; larger windows retain stale mask statistics for longer.

\end{document}